\documentclass[11pt,letterpaper]{article}
\usepackage[margin=1in]{geometry}
\usepackage[T1]{fontenc}
\usepackage{lmodern}
\usepackage{microtype}
\usepackage{amsmath,amssymb}
\usepackage{graphicx}
\usepackage{pdflscape}
\usepackage{booktabs}
\usepackage{array}
\usepackage{tikz}
\usetikzlibrary{arrows.meta,positioning,calc}
\usepackage[hidelinks]{hyperref}
\hypersetup{pdftitle={V-JEPA 2 Meets AQA: Phase-Based Scoring and Reference-Based Feedback for Mountain Biking}}
\newcommand{\R}{\mathbb{R}}

\newcommand{\softmax}{\operatorname{softmax}}
\newcommand{\std}{\operatorname{std}}

\title{V-JEPA 2 Meets AQA: Phase-Based Scoring and Reference-Based Feedback for Mountain Biking}
\author{Santhiya Rajan\textsuperscript{1}\thanks{This project was undertaken by both authors in their personal capacities, outside their professional employment. Affiliations are provided for identification only and do not imply institutional involvement, sponsorship or endorsement.}\quad
Savio Cardoz\textsuperscript{2}\\[0.6em]
\small \textsuperscript{1}Multiverse Computing, Spain\\
\small \textsuperscript{2}ACL Digital, Pune}
\hypersetup{pdfauthor={Santhiya Rajan and Savio Cardoz}}

\date{September 26, 2026}

\begin{document}
\maketitle
\begin{abstract}
We study mountain-bike action quality assessment (AQA) with weak labels and candidate feedback. The training corpus comprises 95 field-recorded jump clips with binary labels denoting good or poor technique. These are coarse categorical labels, not continuous expert quality scores; phase-boundary and aspect-level annotations are unavailable. Our pipeline summarizes four heuristic phases using frozen V-JEPA 2 video features and two-dimensional pose and motion measurements. A small transformer predicts a binary-label score. A separate reference layer compares phase-specific measurements with those of clips labelled as good technique to propose phase, body-parameter, and direction cues. We detail the measurements, network, and evaluation so that the limits of each output are explicit. In five-fold clip-level cross-validation, the full scorer obtained pooled out-of-fold AUROC 0.768 and average precision 0.812. These are exploratory figures because the reporting fold also selected the neural training epoch and riders were not separated between folds. Removing biomechanical scoring inputs raised AUROC to 0.809. A separate inner-selection run obtained AUROC 0.678, but used about 57 rather than 76 training clips per outer fold, so its difference from the test-selected run does not quantify selection bias. The unsupervised aspect head did not learn distinct factors. On a 17-clip external probe, an earlier checkpoint obtained binary accuracy 0.412 and Spearman association $-0.135$ with expert scores. Candidate cues had precision 0.118 and recall 1.0 against each of two coaches, with substantial disagreement about direction. These findings do not establish rider-independent scoring or reliable corrective feedback; they identify the empirical work needed to evaluate either use.
\end{abstract}

\section{Introduction}
\label{sec:introduction}
Action quality assessment estimates how well a movement was performed. Many established datasets contain expert scores for diving, gymnastics, and related activities~\cite{parmar2019whatandhow,tang2020uncertainty,yu2021core}. Field recordings of mountain biking often provide only coarse technique labels. Binary labels denoting good or poor technique do not specify individual technical deficiencies or constitute continuous expert quality assessments. Moreover, viewpoint, terrain, recording equipment, and rider identity can change together with the label.

This study examines one pipeline for mountain-bike jumps under those constraints. A frozen video encoder supplies phase-level appearance and motion features. Measured pose and simple motion descriptors enter the same scorer. A reference layer then compares phase-specific measurements with a distribution fitted to examples labelled as good technique and proposes possible corrective cues. The score and cue generator are evaluated separately. The current evidence does not establish that either output is ready for coaching decisions.

Our central question is what can be learned from the available measurements without treating weak labels as expert quality judgments. We report binary discrimination because the training target has only two values. We also examine three failure points: test-fold epoch selection, lack of rider-disjoint evaluation, and the gap between automatically generated cues and coach judgments. The contribution is an exploratory case study and an explicit accounting of these limits, rather than a new action-quality architecture or a public benchmark.

The quantitative findings are mixed. The full model's reported AUROC is 0.768 under a test-epoch-selected protocol; a variant without biomechanical scoring inputs within the same exploratory comparison reaches 0.809. A separately trained inner-selection arm reaches 0.678, but it uses fewer training examples and cannot isolate the effect of selection. The external probe is adverse for scoring, and feedback precision is low. These outcomes motivate a narrower interpretation of frozen-feature assessment and a concrete evaluation agenda for this domain.

\section{Related Work}
\label{sec:related}
\paragraph{Scoring and skill assessment.}
Action quality assessment has been studied through direct score regression, uncertainty-aware score distributions, exemplar comparisons, and pairwise skill ranking~\cite{parmar2019whatandhow,tang2020uncertainty,yu2021core,doughty2018whosbetter}. AQA-7 examines assessment across multiple actions~\cite{parmar2019aqa7}, whereas FineDiving adds procedure-aware annotations~\cite{xu2022finediving}. Grade-decoupling Likert Transformer combines grade-specific responses into a score~\cite{xu2022gdlt}. TSA-Net aggregates features within tracked spatiotemporal tubes~\cite{wang2021tsa}. Pairwise Contrastive Learning Network learns relative scores between video pairs~\cite{li2022pcln}. These studies provide context for our scorer, but they are not all reproduced as baselines here. Their expert-score or ranking tasks also differ from our binary technique-classification task. Our evaluation cannot be read as continuous quality estimation.

\paragraph{Interpretability and feedback.}
Several closer studies address low-data adaptation or feedback. PECoP studies parameter-efficient continual pretraining for action quality assessment~\cite{dadashzadeh2024pecop}. IRIS and Poze use rubric-based segmentation and pose-based sports feedback, respectively~\cite{matsuyama2023iris,singh2024poze}. ExpertAF studies actionable expert feedback from video~\cite{ashutosh2025expertaf}. Hierarchical neuro-symbolic assessment connects measured motion to explanations~\cite{okamoto2024hierarchical}. Narrative Action Evaluation generates evaluative commentary through prompt-guided multimodal interaction~\cite{zhang2024nae}. FineCausal introduces causal-intervention and temporal-attention modules for fine-grained assessment~\cite{han2025finecausal}. Ego-Exo4D provides synchronized first- and third-person skilled-activity recordings with expert commentary~\cite{grauman2024egoexo4d}. These richer annotations differ from our binary technique labels and small coach-reviewed probe. Our combination of a frozen encoder, a small scorer, and reference-based cue rules is not presented as a technical advance over these systems. The distinct setting here is weakly labelled field footage of mountain-bike jumps and a direct audit of what its available evaluation supports.

\paragraph{Video and language representations.}
VideoMAE learns from masked video reconstruction~\cite{tong2022videomae}; InternVideo combines generative and discriminative video-learning objectives~\cite{wang2022internvideo}. The original V-JEPA learns by predicting video features~\cite{bardes2024vjepa}. V-JEPA 2, rather than the original V-JEPA, is the frozen encoder used in this study~\cite{assran2025vjepa2}. LLaVA and the original InternVL study visual instruction following and vision--language alignment~\cite{liu2024llava,chen2024internvl}. The original Gemma report concerns language models, not a video-coaching validator~\cite{gemmateam2024gemma}. These are background studies, not additional components of our system. Our optional wording-and-direction step uses InternVL3~\cite{zhu2025internvlthree}; its outputs are not independent evidence of cue correctness.

\paragraph{Pose and biomechanical measurement.}
YOLO-Pose jointly detects people and 2D keypoints~\cite{maji2022yolopose}. It is cited as related pose-estimation work, not as the exact YOLOv8 checkpoint used here. VideoPose3D lifts 2D keypoint sequences into 3D~\cite{pavllo2019videopose3d}, ST-GCN models skeleton dynamics for action recognition~\cite{yan2018stgcn}, and MotionBERT learns motion representations through 3D-motion recovery from partial 2D observations~\cite{zhu2023motionbert}. Physics-informed posture refinement additionally imposes anatomical constraints~\cite{leuthold2025physics}. None of these 3D or skeleton-learning methods is part of the evaluated pipeline. Our descriptors remain camera-relative 2D measurements; they do not establish view-invariant kinematics.

\paragraph{Cycling context and limits of transfer.}
OpenCap estimates movement kinematics and dynamics from synchronized smartphone views~\cite{uhlrich2023opencap}; Kakavand et al. compare it with marker-based measurements during cycling~\cite{kakavand2024opencap}. Video-based cyclocross analysis instead tracks riding lines and behavior~\cite{debock2021cyclocross}. Cycling biomechanics covers joint motion, bicycle configuration, and rider posture~\cite{bini2014cycling}, while a review of uphill cycling discusses changes associated with slope and riding position~\cite{fonda2012uphill}. The BikeFit IA product website describes video-based cycling-position analysis~\cite{bikefitia}; this is a product reference, not peer-reviewed validation of coaching accuracy. Engeler et al. study slope-dependent posture in running, a different activity whose findings do not supply mountain-bike jump targets~\cite{engeler2025downhill}. These sources motivate attention to measurement context, but none validates our numerical reference thresholds. An inspectable measurement is not evidence that the proposed correction is biomechanically appropriate. Viewpoint, occlusion, and erroneous phase boundaries remain central concerns in the external feedback assessment below.

\section{Jump Specification and Available Supervision}
\label{sec:specification}
The jump configuration contains four ordered phase names, 13 measurable parameters, and eight named aspects. It specifies which parameters a cue may use for each aspect and which phase name to attach to that cue. This configuration is a structured description of the task; it is not a set of observed expert aspect scores. Only jumps were evaluated. The ability to write a different configuration does not establish transfer to another technique.

\begin{table}[t]
\centering
\small
\begin{tabular}{p{0.27\linewidth}p{0.16\linewidth}p{0.47\linewidth}}
\toprule
Aspect & Assigned phase & Parameters eligible for its cue \\
\midrule
Approach speed control & Approach & Residual image-motion magnitude \\
Take-off extension & Take-off & Knee angle, hip angle, launch-angle proxy \\
Take-off arm position & Take-off & Elbow angle \\
Flight body--bike separation & Flight & Vertical body--bike offset, hip angle \\
Flight balance & Flight & Horizontal body--bike offset, torso orientation \\
Airborne control & Flight & Quadratic-fit residual, rotation-variation descriptor, apex-height proxy \\
Landing absorption & Landing & Knee angle, elbow angle \\
Landing balance & Landing & Horizontal body--bike offset, torso orientation \\
\bottomrule
\end{tabular}
\caption{The eight aspect-to-phase and aspect-to-parameter assignments used for jumps. The phase column is a predetermined association in the cue generator. It must not be interpreted as a learned phase prediction.}
\label{tab:aspect_spec}
\end{table}

Table~\ref{tab:parameters} lists all feature parameters. The three joint angles use the right-side shoulder, elbow, wrist, hip, knee, and ankle keypoints according to the corresponding joint triple. Torso orientation uses the vector between mid-hip and mid-shoulder. The horizontal and vertical offset compare a four-keypoint body-center proxy with a detected bike-box center and divide by image width and height, respectively. These definitions constrain the meaning of a reported cue: for example, ``balance'' in this work is an image-plane offset, not a measured mechanical center of mass.

\begin{table}[t]
\centering
\small
\begin{tabular}{p{0.34\linewidth}p{0.14\linewidth}p{0.42\linewidth}}
\toprule
Parameter & Unit & Measurement used \\
\midrule
Elbow, knee, hip angle & degrees & Right-side 2D keypoint triples \\
Torso orientation & degrees & Mid-hip to mid-shoulder inclination from image vertical \\
Horizontal, vertical body--bike offset & image ratio & Body-center proxy minus bike-box center, divided by frame size \\
``Speed'' feature & pixels/frame & Mean residual dense-flow magnitude after global motion compensation \\
``Launch angle'' & degrees & Image-plane body-center displacement around start of assigned flight window \\
``Apex height'' & image ratio & First-to-minimum body-center image height within assigned flight window \\
``Airtime fraction'' & fraction & Assigned flight-window length divided by clip length \\
Ballistic residual & unitless & Quadratic-fit root-mean-square residual divided by trajectory range \\
Rotation rate & degrees/frame & Mean absolute change in torso orientation \\
``Rotation smoothness'' & degrees/frame & Standard deviation of frame-to-frame torso-orientation change \\
\bottomrule
\end{tabular}
\caption{Seven pose or motion parameters and six flight descriptors. Quoted names are the system's feature labels; all measurements are image-relative proxies. The final descriptor is smaller when observed rotation is more uniform.}
\label{tab:parameters}
\end{table}

The binary technique label is encoded as $y=0.85$ for good technique and $y=0.30$ for poor technique on the unit scale. These values encode two classes; their numeric gap is not an expert quality interval. The corpus has no labels for the eight aspect scores and no cue labels for training the deficiency-output head. The 17-clip external probe described in Section~\ref{sec:external} was annotated separately by two coaches for evaluation. Table~\ref{tab:supervision} makes the distinction explicit.

\begin{table}[t]
\centering
\small
\begin{tabular}{p{0.32\linewidth}p{0.23\linewidth}p{0.34\linewidth}}
\toprule
Field & Main 95 clips & External 17 clips \\
\midrule
Binary technique class & 52 good technique; 43 poor technique & Available \\
Continuous expert score & Not available & Two-coach scores used in the external probe \\
Aspect-level score labels & Not available & Not used to train the scorer \\
Corrective-cue annotations & Not available for training & Two coaches; 136 clip--aspect decisions \\
Independent phase boundaries & Not available & Not available for temporal localization testing \\
Rider/session/site identifiers & Not recorded & Insufficient for a rider-disjoint test of the main model \\
\bottomrule
\end{tabular}
\caption{What is observed and what is only generated by the pipeline. The coach-annotated probe evaluates an earlier checkpoint, not the main comparison model.}
\label{tab:supervision}
\end{table}

\section{System}
\label{sec:method}
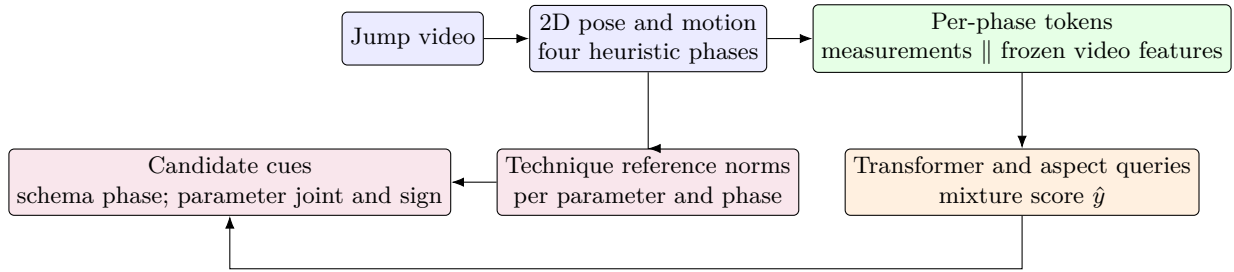
\begin{figure}[t]
\centering
\begin{tikzpicture}[>=Latex, node distance=5mm, every node/.style={font=\footnotesize},
 box/.style={draw,rounded corners=2pt,align=center,minimum height=7mm,inner sep=3pt},
 blue/.style={box,fill=blue!8}, green/.style={box,fill=green!10},
 orange/.style={box,fill=orange!12}, purple/.style={box,fill=purple!10}]
\node[blue] (video) {Jump video};
\node[blue,right=6mm of video] (phase) {2D pose and motion\\four heuristic phases};
\node[green,right=6mm of phase] (tokens) {Per-phase tokens\\measurements $\Vert$ frozen video features};
\node[orange,below=10mm of tokens] (model) {Transformer and aspect queries\\mixture score $\hat y$};
\node[purple,left=6mm of model] (norms) {Technique reference norms\\per parameter and phase};
\node[purple,left=6mm of norms] (cues) {Candidate cues\\schema phase; parameter joint and sign};
\draw[->] (video) -- (phase);
\draw[->] (phase) -- (tokens);
\draw[->] (tokens) -- (model);
\draw[->] (phase.south) |- (norms.north);
\draw[->] (model.south) -- ++(0,-7mm) -| (cues.south);
\draw[->] (norms) -- (cues);
\end{tikzpicture}
\caption{Schematic information flow. The full score consumes video and biomechanical features. Candidate cues also use the scorer's aspect outputs and a separate parameter comparison. The cue phase is assigned by the aspect schema, rather than localized by learned attention. No branch in this diagram is claimed to provide validated interpretation of the score.}
\label{fig:system}
\end{figure}

The system has two outputs: a clip-level score for the binary technique class and a list of candidate corrective cues. Figure~\ref{fig:system} shows the information paths. Pose and motion measurements are part of the \emph{full scorer's input} as well as the separate reference comparison used to construct cues. These roles must be distinguished when interpreting the ablations.

\subsection{From a field clip to four phase windows}
\label{sec:preprocessing}
Frames are decoded at their native rate. Frozen 2D pose estimation supplies 17 body keypoints per frame; the evaluated configuration used a small YOLOv8 pose model and did not load a 3D lifter. A second frozen detector selects the highest-confidence bicycle or motorcycle box above 0.30 confidence. The pose measurement and bike-box center are used as inputs to the parameter definitions in Table~\ref{tab:parameters}. Failure to detect either can produce zero-valued feature entries, so a numeric zero need not mean an observed angle or offset of zero. The body-center proxy is the mean of the two shoulder and two hip keypoints. It is a projected point, not the physical center of mass.

For an available skeleton, the pose geometry subtracts the midpoint of the hips $\mathbf c_h$ and divides by the distance from that point to the midpoint of the shoulders $\mathbf c_s$:
\begin{equation}
\widetilde{\mathbf j}_n=
\frac{\mathbf j_n-\mathbf c_h}{\|\mathbf c_s-\mathbf c_h\|+\epsilon}.
\label{eq:pose_norm}
\end{equation}
This removes image translation and a scale factor; it does not remove perspective or camera viewpoint. The angle at keypoint $b$ between neighbors $a$ and $c$ is computed in degrees as
\begin{equation}
\theta_b=\frac{180}{\pi}\arccos\!\left[
\operatorname{clip}\!\left(
\frac{(\mathbf j_a-\mathbf j_b)^{\mathsf T}(\mathbf j_c-\mathbf j_b)}
{\|\mathbf j_a-\mathbf j_b\|\,\|\mathbf j_c-\mathbf j_b\|+\epsilon},-1,1\right)\right].
\label{eq:joint_angle}
\end{equation}
Torso orientation is the absolute inclination of the mid-hip--mid-shoulder vector from the image vertical. The detected bike box defines a center $\mathbf c_b$; the normalized body--bike offset is $((c_{r,x}-c_{b,x})/W,(c_{r,y}-c_{b,y})/H)$ for frame width $W$ and height $H$. This is a geometric image comparison rather than a statement about rider weight distribution.

To form the feature labelled ``speed,'' dense frame-to-frame flow is corrected by a global homography estimated from ORB feature matches. If $\mathbf u_t(\mathbf p)$ is dense flow and $H_t$ the fitted homography, the residual is
\begin{equation}
\mathbf u_t^{\rm res}(\mathbf p)=\mathbf u_t(\mathbf p)-
\left(\operatorname{proj}(H_t\widetilde{\mathbf p})-\mathbf p\right),
\qquad r_t=\frac{1}{WH}\sum_{\mathbf p}\|\mathbf u_t^{\rm res}(\mathbf p)\|_2,
\label{eq:flow}
\end{equation}
where $\widetilde{\mathbf p}$ is the homogeneous pixel coordinate and $r_t$ is the frame-wide mean magnitude. If sufficient matches are unavailable, the uncompensated flow magnitude is used. A global homography cannot remove parallax or isolate the rider from other moving image regions; $r_t$ is a residual scene-motion proxy in pixels per frame, not calibrated rider speed.

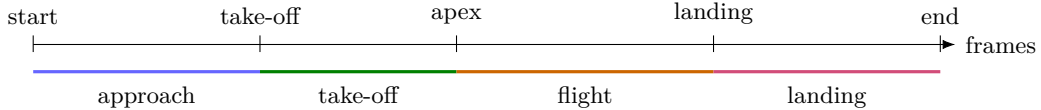
\begin{figure}[t]
\centering
\begin{tikzpicture}[x=1cm,y=1cm,>=Latex,font=\footnotesize]
\draw[->] (0,0) -- (12.2,0) node[right] {frames};
\foreach \x/\name in {0/start,3/take-off,5.6/apex,9/landing,12/end}{
  \draw (\x,-0.10)--(\x,0.12);
  \node[above,align=center] at (\x,0.14) {\name};
}
\draw[very thick,blue!60] (0,-0.36)--(3,-0.36);
\draw[very thick,green!50!black] (3,-0.36)--(5.6,-0.36);
\draw[very thick,orange!80!black] (5.6,-0.36)--(9,-0.36);
\draw[very thick,purple!70] (9,-0.36)--(12,-0.36);
\node[below] at (1.5,-0.42) {approach};
\node[below] at (4.3,-0.42) {take-off};
\node[below] at (7.3,-0.42) {flight};
\node[below] at (10.5,-0.42) {landing};
\end{tikzpicture}
\caption{Operational phase partition. The three interior boundaries come from image-plane hip motion and are constrained to occur in order. The widths are schematic. The assigned flight window begins at the estimated apex; the labels have no independently measured temporal accuracy.}
\label{fig:phase_timeline}
\end{figure}

The phase heuristic smooths the vertical hip trajectory $y_t$ with a short moving average and forms first differences $v_t=y_t-y_{t-1}$. Image $y$ increases downward. It chooses the strongest upward change as take-off, the minimum subsequent image height as the apex, and the strongest subsequent downward change as landing, subject to coarse constraints on their positions in the clip. The four windows span start--take-off, take-off--apex, apex--landing, and landing--end, labelled approach, take-off, flight, and landing (Figure~\ref{fig:phase_timeline}). For clips with fewer than twice as many frames as phases, it uses equal-length splits. The phase names are therefore operational labels for windows, not independently verified event boundaries.

\subsection{Phase features and scoring}
\label{sec:scoring}
The four windows defined in Section~\ref{sec:preprocessing} serve as pooling regions for the seven per-frame measurements and the frozen video features. Six additional flight descriptors occupy selected phase slots. Table~\ref{tab:aspect_spec} specifies which parameters are relevant to each proposed cue; the scorer itself receives every parameter in every phase token.

For each non-flight parameter $k$ in phase $p$, the feature assembly records the within-window mean $\mu_{kp}$, standard deviation $\sigma_{kp}$, and change $\delta_{kp}$. Six flight descriptors $f_{p\ell}$ are stored as $\boldsymbol\pi_p=[(f_{p\ell},0,0)]_{\ell=1}^{6}$ to share the same feature layout. The video branch samples 16 RGB frames evenly from each phase (repeating frames for shorter windows), passes them through the pretrained model's video processor, and runs a frozen V-JEPA 2 ViT-L encoder~\cite{assran2025vjepa2}. The processor handles spatial resizing and normalization; its resolved numeric settings were not retained with the results. Hidden tokens are averaged to a 1,024-dimensional vector $\mathbf v_p$. The assembled token is
\begin{equation}
\mathbf{x}_p=\big[\{\mu_{kp},\sigma_{kp},\delta_{kp}\}_{k\in\mathcal K_{\rm nonflight}}\;\Vert\;\boldsymbol\pi_p\;\Vert\;\mathbf{v}_p\big]\in\R^{1063},
\label{eq:token}
\end{equation}
with seven non-flight parameters contributing $7\times3=21$ entries, six flight descriptors contributing $6\times3=18$, and frozen video features contributing 1,024. The scorer receives $\mathbf X=[\mathbf x_1;\ldots;\mathbf x_4]\in\R^{4\times1063}$. The measurements are camera-relative proxies, not calibrated estimates of physical speed or center of mass.

For clarity, the flight descriptors are defined from the assigned flight window $[s,e)$ and the image-plane body-center trajectory $(x_t,y_t)$, with $N_a=e-s$ and total clip length $T$. A least-squares quadratic $\widehat y_t=at^2+bt+c$ gives
\begin{equation}
r_{\rm quad}=\frac{\sqrt{N_a^{-1}\sum_{t=s}^{e-1}(y_t-\widehat y_t)^2}}
{\max_{s\le t<e}y_t-\min_{s\le t<e}y_t+\epsilon},\qquad
h_{\rm win}=\frac{y_s-\min_{s\le t<e}y_t}{H},\qquad
a_{\rm win}=\frac{N_a}{T}.
\label{eq:flight}
\end{equation}
The so-called launch-angle feature uses the mean displacement in a five-frame window around $s$, the \emph{start of the assigned flight window}, and is stored in the take-off feature slot:
\begin{equation}
\theta_{\rm win}=\frac{180}{\pi}\operatorname{atan2}(-\overline{\Delta y},|\overline{\Delta x}|+\epsilon).
\label{eq:launch}
\end{equation}
Because the phase heuristic places $s$ at the estimated apex, this value is not a verified physical launch angle. Likewise $h_{\rm win}$ is measured from the first frame of the assigned flight window rather than from the actual take-off. The terms name stored features, not independently validated jump events.

The two orientation descriptors are $\overline{|\Delta\phi|}$ and
\begin{equation}
q_{\rm rot}=\std_{t}\!\left(\phi_{t+1}-\phi_t\right),
\label{eq:rotation}
\end{equation}
where $\phi_t$ is torso orientation in degrees and $\Delta\phi=\phi_{t+1}-\phi_t$. The feature called ``rotation smoothness'' is the variability of frame-to-frame orientation change; lower values indicate more uniform observed rotation. It does not use angular acceleration or an exponential transform. All six descriptors remain sensitive to camera motion, perspective, and segmentation error.

\begin{figure}[t]
\centering
\begin{tikzpicture}[>=Latex,font=\footnotesize,node distance=5mm and 6mm,
 b/.style={draw,rounded corners=2pt,align=center,inner sep=3pt,minimum height=8mm}]
\node[b,fill=green!9] (x) {Four tokens\\$4\times1063$};
\node[b,fill=orange!9,right=of x] (proj) {Projection + phase positions\\$4\times128$};
\node[b,fill=orange!9,right=of proj] (enc) {Two encoder layers\\$4\times128$};
\node[b,fill=blue!9,below=10mm of proj] (query) {Eight queries\\$8\times128$};
\node[b,fill=orange!9,below=10mm of enc] (attn) {Cross-attention\\$8\times128$; weights $8\times4$};
\node[b,fill=purple!9,below=12mm of attn] (heads) {Score + deficiency heads\\eight values each};
\node[b,fill=purple!9,left=of heads] (mix) {Softmax mixture\\one clip score};
\draw[->] (x)--(proj);
\draw[->] (proj)--(enc);
\draw[->] (enc)--(attn);
\draw[->] (query)--(attn);
\draw[->] (attn)--(heads);
\draw[->] (heads)--(mix);
\end{tikzpicture}
\caption{Trainable head for one clip (batch dimension suppressed). The mixture uses only score-head outputs. Only the clip-level mixture score has an observed training target here. Aspect scores, deficiency outputs, and phase-attention weights have no matching supervision and are not validated explanations.}
\label{fig:head}
\end{figure}
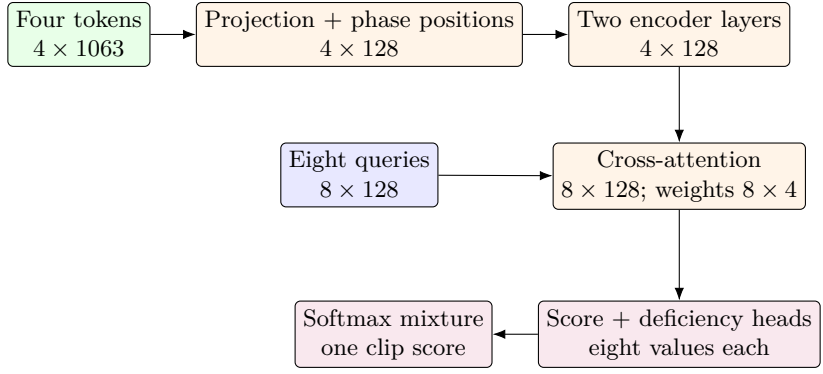

A two-layer transformer (width 128, four attention heads) processes the four tokens with phase embeddings. Eight learned aspect queries attend to its outputs. A sigmoid head returns each aspect sub-score $\hat s_j$, and the clip score is a convex mixture,
\begin{equation}
\hat y=\sum_{j=1}^{8} w_j\hat s_j,
\qquad \mathbf{w}=\softmax(\boldsymbol\theta).
\label{eq:score}
\end{equation}
The head has 485,898 trainable parameters, allocated by module in Table~\ref{tab:head_params}.
The input projection maps $1063\rightarrow128$ and adds one learned 128-dimensional embedding for each of the four phase positions. Two transformer-encoder layers have four attention heads and feed-forward width 256. A bank of eight learned 128-dimensional queries cross-attends to the resulting four phase vectors. The eight attended vectors each pass through a shared $128\rightarrow64\rightarrow1$ score head and a separate deficiency-output head. The score head has a sigmoid output; the eight score values are combined by Eq.~\ref{eq:score}. A cross-attention matrix has shape $8\times4$, but it is not trained against phase labels. Figure~\ref{fig:head} records the dimensions and the status of each output.

\begin{table}[t]
\centering
\small
\begin{tabular}{lr}
\toprule
Trainable module & Parameters \\
\midrule
Input projection and normalization & 136,448 \\
Four phase embeddings & 512 \\
Two transformer encoder layers & 264,960 \\
Eight aspect queries & 1,024 \\
Cross-attention & 66,048 \\
Aspect normalization & 256 \\
Score head & 8,321 \\
Deficiency-output head & 8,321 \\
Eight aggregation logits & 8 \\
\midrule
Total & 485,898 \\
\bottomrule
\end{tabular}
\caption{Trainable parameter count for the jump head. The pose, bike, and V-JEPA 2 backbones were frozen. The deficiency-output head is included in the count despite having no cue-level training target.}
\label{tab:head_params}
\end{table}

There are no aspect labels in the 95-clip training corpus. Training therefore uses only the overall binary technique target. With $y_i\in\{0.30,0.85\}$ denoting the two class encodings, the objective is
\begin{equation}
\mathcal L=\operatorname{MSE}(\hat y,y)+0.3\,\mathcal L_{\rm rank},\quad
\mathcal L_{\rm rank}=\frac{1}{|\mathcal Q|}\sum_{(i,j)\in\mathcal Q}
\max\{0,\,0.05-\operatorname{sgn}(y_i-y_j)(\hat y_i-\hat y_j)\},
\label{eq:loss}
\end{equation}
where $\mathcal Q$ contains within-batch pairs with different labels; the ranking term is zero if no such pair is present. The head uses dropout 0.4, AdamW with learning rate $3\times10^{-4}$ and weight decay $10^{-3}$, batch size eight, cosine learning-rate annealing, gradient clipping at norm 1, and at most 120 epochs with patience 25. The epoch-selection criterion is validation-fold Spearman association for the overall label plus the mean aspect association, which contributes no labelled signal here. Using the reporting fold for that choice is the principal evaluation defect discussed in Section~\ref{sec:data}. Seed 42 initializes the evaluation, with fold-indexed seeds for neural fitting. Training examples receive Gaussian feature noise of standard deviation 0.03; two additional augmented views per clip use multiplicative jitter of at most 4\% and additive noise of standard deviation 0.05. Evaluation features are unaugmented. Aspect scores and attention are model internals rather than supervised quality or phase predictions; Section~\ref{sec:results} tests whether they specialize.

\subsection{Reference comparison and candidate cues}
\label{sec:cues}
For each measured parameter and phase, a reference mean $\mu^*_{kp}$ and standard deviation $\sigma^*_{kp}$ are fitted from clips labelled as good technique. This reference is conditioned on the positive technique class, not on independently adjudicated expert ratings. A new measurement is standardized as
\begin{equation}
z_{kp}=\frac{\mu_{kp}-\mu^*_{kp}}{\sigma^*_{kp}+\epsilon}.
\label{eq:z}
\end{equation}
Within each aspect, the generator chooses the associated parameter with the largest $|z|$. It emits a cue if its aspect score is below 0.6, its deficiency output exceeds 0.5, \emph{or} $|z|\geq1.5$. The deficiency-output head received no cue-level supervision in this corpus, so its trigger is not a trained defect detector. The cue's phase comes from the aspect specification. Its joint or body parameter comes from the selected parameter, and its ``too low'' or ``too high'' direction is the sign of $z$. Severity combines sub-score and deviation magnitude. A reference deviation establishes that the measured value differs from this sample's positive-class reference distribution; it does not establish that changing it in the proposed direction would improve a jump.

The reference norms were fitted on the full labelled corpus for the separately evaluated feedback layer. They do not enter the scalar scoring calculation, but this fitting choice prevents the cue results from being interpreted as held-out reference evaluation within the 95 clips. For the external probe, both the checkpoint and stored norms came from an earlier run than the one summarized in the main comparison. An optional vision--language wording and direction module was examined separately; it is disabled for the reported primary cues.

\begin{figure}[t]
\centering
\begin{tikzpicture}[>=Latex,font=\footnotesize,node distance=6mm and 7mm,
 b/.style={draw,rounded corners=2pt,align=center,inner sep=3pt,minimum height=8mm}]
\node[b,fill=green!9] (measure) at (0,1.6) {Phase measurement\\$\mu_{kp}$};
\node[b,fill=green!9] (ref) at (0,-.1) {Technique reference\\$\mu^*_{kp},\sigma^*_{kp}$};
\node[b,fill=orange!9] (z) at (4.0,.75) {Standardized deviation\\$z_{kp}$; select largest $|z|$};
\node[b,fill=purple!9] (score) at (9.1,2.2) {Aspect score + deficiency\\outputs from scorer};
\node[b,fill=orange!9] (gate) at (9.1,.75) {OR gate\\score, deficiency, deviation};
\node[b,fill=purple!9] (cue) at (9.1,-1.25) {Candidate cue\\schema phase, parameter\\sign, severity};
\node[b,fill=blue!7] (vlm) at (4.0,-1.25) {Optional InternVL3\\four phase frames\\+ measurements};
\draw[->] (measure)--(z);
\draw[->] (ref)--(z);
\draw[->] (z)--(gate);
\draw[->] (score)--(gate);
\draw[->] (gate)--(cue);
\draw[->,dashed] (cue)--(vlm);
\draw[->,dashed] (vlm.south) -- ++(0,-5mm) -| (cue.south);
\end{tikzpicture}
\caption{Candidate-cue dataflow. The scorer supplies aspect and deficiency outputs to the OR gate, although the deficiency head has no cue-level training target; measurements and positive-class reference statistics supply the deviation. The optional vision--language step can alter direction and wording for cues with a keypoint-linked parameter, but does not change the score, chosen parameter, or schema phase. Dashed arrows denote an optional path with no established benefit.}
\label{fig:feedback_flow}
\end{figure}
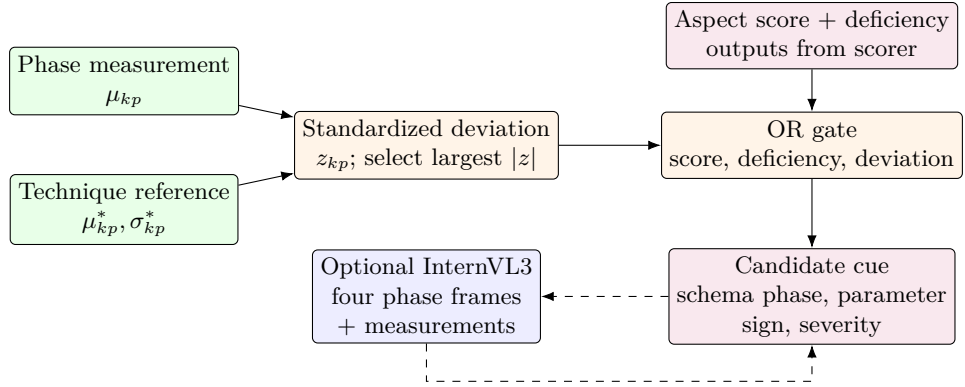

Figure~\ref{fig:feedback_flow} separates the scorer outputs, reference deviation, cue gate, and optional wording stage.
\paragraph{Candidate construction and scope.}
The reference set uses the upper 60th-percentile threshold of the two-valued technique encoding. For the observed class counts this threshold equals the good-technique target value, so all training clips labelled as good technique are eligible. Means and standard deviations are calculated separately for each parameter and phase from the available measured values; missing reference combinations fall back to a neutral default. For an aspect assigned to phase $p$, the rule selects $k^*=\arg\max_{k\in\mathcal K_j}|z_{kp}|$. The phase label in the cue is the aspect's configured phase; the joint label is a human-readable name attached to $k^*$. If $z_{k^*p}<0$, the direction field says ``too low''; otherwise it says ``too high.'' These words describe deviation from the estimated reference mean, not a counterfactual improvement.

The severity calculation is also deterministic. Writing $s_j$ for the aspect score, it forms $r_j=\max\{1-s_j,\min(|z_{k^*p}|/3,1)\}$ and assigns low, medium, or high for $r_j<0.33$, $0.33\leq r_j<0.66$, or $r_j\geq0.66$. At most one cue is produced per aspect, so a clip can receive up to eight. The generator sorts emitted cues by severity and score. The OR gate described above means a cue can be emitted even when $|z|<1.5$ if a score or deficiency output triggers it. This behavior helps explain why the external assessment found high recall and low precision; it does not identify which trigger caused each error.

When enabled, the optional InternVL3 step~\cite{zhu2025internvlthree} receives up to four frames from the assigned phase, the selected body-parameter name, the measured and reference values, and the signed deviation. It proposes a direction and a short sentence. Missing frames, unavailable model, malformed reply, or empty message retain the original cue; an invalid direction also falls back to the sign rule. The option is off for the main feedback evaluation. The 20-cue examination in Section~\ref{sec:external} measures how often this path changes a direction, not whether it makes a correct intervention.

\section{Data and Evaluation Protocol}
\label{sec:data}
The main corpus contains 95 jump clips: 52 labelled as good technique and 43 labelled as poor technique (Table~\ref{tab:dataset}). Mean duration is 3.74 seconds (standard deviation 4.33; range 0.71--42.5). All recordings have a resolution of 1280$\times$720 pixels, at approximately 24, 25, 30, or 60 frames per second. The training targets numerically encode the two technique classes. They are not per-clip expert quality scores. The main corpus has no aspect, cue, or phase-boundary labels. Coaches reviewed a limited subset of the technique assignments. However, the retained records do not specify the subset size or inter-rater agreement sufficiently to quantify label reliability.

\begin{table}[t]
\centering
\small
\begin{tabular}{ll}
\toprule
Property & Observed main-corpus value \\
\midrule
Jump clips & 95 (52 good technique, 43 poor technique) \\
Duration & $3.74\pm4.33$ seconds; range 0.71--42.5 \\
Frame rates & 23.98, 25.00, 29.97, 59.94 frames/s \\
Resolution & $1280\times720$ pixels for all clips \\
Training target & 0.85: good technique; 0.30: poor technique \\
Aspect, cue, phase-boundary labels & None \\
Rider, location, camera identifiers & Not recorded \\
Main folds & Five; one unique group per clip \\
\bottomrule
\end{tabular}
\caption{Main corpus description. The two target values are encodings of technique categories. They do not represent expert-rated scores or a 0.55-unit difference in technique quality.}
\label{tab:dataset}
\end{table}

One training example contains the clip, the jump identity, and the overall technique-class target. No aspect-level or corrective-cue training targets are provided. This is important for interpreting the head: eight sub-scores and eight deficiency outputs are produced at inference, but none was trained against a corresponding expert aspect or cue target. The separate coach annotations belong only to the 17-clip probe; they were not used to fit the main model.

Rider, location, session, and camera identifiers were not recorded. The five-fold grouping procedure therefore assigns each clip its own group. It prevents reuse of the same clip across training and test folds, but it does not separate riders, recording sessions, or trails. Similar scene or acquisition conditions can occur in both partitions. All scalar and adapted-baseline comparisons below use one seed and those clip-level folds. The reported values are exploratory in-corpus discrimination, not estimates of performance on new riders or locations.

The principal binary metrics are area under the receiver operating characteristic curve (AUROC) and average precision (AP), computed from pooled out-of-fold predictions. Good technique is the positive class; its prevalence is $52/95\simeq0.547$. Pooling scores from separately fitted folds assumes comparable score scales, which has not been checked. We therefore use these metrics descriptively. Spearman correlation with the two-valued target is a tie-limited association measure, not evidence that the model ranks continuous technique quality. No inferential comparison between methods is claimed.

For the proposed scorer, its ablations, and the MLP, GRU, and mean-pool transformer baselines, the reporting fold was also used to choose the training epoch. The adapted CoRe, USDL, and pairwise heads were fit for a fixed 120 epochs, and the classical estimators have no training epoch to select. Test-fold selection can favor the first group's reported results. A second diagnostic for the proposed scorer used the same outer fold allocation but selected the epoch on an inner validation split. In a typical outer fold the first arm trained on 76 clips; the second trained on roughly 57 and did not retrain on all 76 after selection. The two arms consequently change training-set size as well as selection access. Neither their difference nor the inner-selection value is a corrected estimate of generalization, and the inner-selection arm has no matched baseline suite.

An additional probe contains 17 clips with technique labels and expert scores. It was assessed with an earlier saved scorer and its associated reference norms, not the model fitted for the main comparison. It is therefore an external diagnostic of a related checkpoint, not a direct external validation of the full model in Table~\ref{tab:models}. Two coaches also annotated corrective cues for these clips. Cue precision and recall count matches at the clip--aspect level; phase, joint, and direction accuracy are conditional on matched cues. The inter-annotator agreement calculation covers 136 assessed items.

\subsection{What the two epoch-selection arms compare}
\label{sec:protocol_detail}
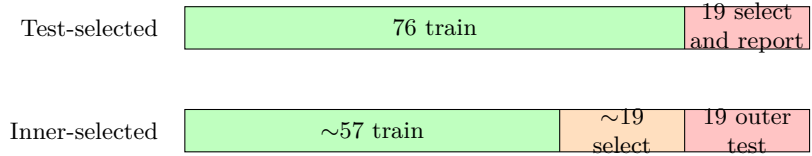
\begin{figure}[t]
\centering
\begin{tikzpicture}[x=0.87cm,y=0.8cm,>=Latex,font=\footnotesize]
\node[anchor=east] at (0,1.8) {Test-selected};
\node[anchor=east] at (0,0.1) {Inner-selected};
\fill[green!25] (0.25,1.45) rectangle (7.85,2.15);
\fill[red!23] (7.85,1.45) rectangle (9.75,2.15);
\draw (0.25,1.45) rectangle (9.75,2.15);
\draw (7.85,1.45)--(7.85,2.15);
\node at (4.05,1.8) {76 train};
\node[align=center] at (8.8,1.8) {19 select\\and report};
\fill[green!25] (0.25,-0.25) rectangle (5.95,0.45);
\fill[orange!25] (5.95,-0.25) rectangle (7.85,0.45);
\fill[red!23] (7.85,-0.25) rectangle (9.75,0.45);
\draw (0.25,-0.25) rectangle (9.75,0.45);
\draw (5.95,-0.25)--(5.95,0.45);
\draw (7.85,-0.25)--(7.85,0.45);
\node at (3.1,0.1) {$\sim$57 train};
\node[align=center] at (6.9,0.1) {$\sim$19\\select};
\node[align=center] at (8.8,0.1) {19 outer\\test};
\end{tikzpicture}
\caption{One outer-fold comparison, with colors denoting training (green), epoch selection (orange or red), and outer reporting (red). The diagnostic arms differ in both data access and fitting-set size. The widths illustrate counts, not a statistical confidence interval.}
\label{fig:protocol}
\end{figure}

Figure~\ref{fig:protocol} shows why the two diagnostic AUROC values cannot be subtracted to obtain a selection-bias estimate. Both use five outer folds of 19 clips each. The test-selected arm trains on the other 76 clips and chooses its epoch using the 19 clips it reports. In the inner-selection arm, those 76 clips are split again: approximately 57 train the model and 19 select its epoch. That fitted model is then evaluated on the untouched outer 19. It is not retrained on 76 clips after epoch selection. The second arm removes direct use of the outer test fold for epoch selection but also removes one quarter of the available outer-training clips from fitting. It was run for the proposed model only. The main baseline table is from another test-selected run, so its full-model AUROC 0.768 is not the same fitted run as the diagnostic arm's 0.781.

\subsection{Metrics and annotation matching}
With good technique as the positive class, AUROC is the probability that a randomly chosen clip labelled as good technique receives a higher score than a clip labelled as poor technique, counting ties at one half. AP summarizes precision across score thresholds and depends on the positive-class prevalence of $52/95$. Reported main values pool predictions from the five fitted fold models. Such pooling can mix fold-specific score scales, so we also retain the original fold-level accuracy and rank association as descriptive context where available. The binary decision threshold for reported accuracy is 0.5. These metrics test association with technique labels, not coaching quality.

For the external feedback probe, a candidate is counted correct for precision and recall when its clip and named aspect match a coach annotation. Each annotated clip--aspect pair is counted once. Joint, direction, and phase comparisons are then made only for matched candidates; in this sample there were 16 such matches per coach. Cohen's $\kappa$ compares whether each coach flagged each of eight aspects in each of 17 shared clips, for $17\times8=136$ binary decisions. It is not a measure of agreement on the full text of the correction. This matching protocol and the schema-assigned phase mean that a high phase figure alone cannot validate temporal localization.

\subsection{Adaptation of comparison methods}
The classical controls receive four-phase features after averaging over phase. Ridge regression uses regularization strength 1, the radial-basis support-vector regressor uses $C=2$, the random forest uses 300 trees, and gradient boosting uses its library defaults with the run seed. Separate ridge probes retain only the pooled frozen-video dimensions or only the pooled measured dimensions. The MLP, GRU, and mean-pool transformer receive the full four-token sequence and use the same overall-score and ranking objective as the proposed head; their selected training epochs have the same reporting-fold access problem.

The three prior AQA comparisons are small adaptations to the binary labels and pooled video representation, not retraining of their original complete systems. Pairwise ranking learns a logistic preference for a clip labelled as good technique over a clip labelled as poor technique. The USDL adaptation maps the two class encodings to Gaussian soft targets over 20 score bins and reads the expected score. The CoRe adaptation predicts score differences relative to a balanced bank of training-fold exemplars. These heads use training-fold feature standardization, two augmented feature views, and a fixed 120-epoch schedule. Their comparison with the proposed head changes representation, temporal pooling, and objective together, so the table is descriptive rather than a controlled method ranking.

\section{Input, Output, and Worked Examples}
\label{sec:examples}
The system accepts a complete jump video and returns a clip-level score,
eight aspect sub-scores, and a variable-length list of candidate cues.
The visual output places the inferred phase and selected cue text above the
video, overlays the detected skeleton, and highlights the parameter selected
by the reference comparison. These are displays of model outputs, not expert
annotations. A score in $[0,1]$ is also not a calibrated probability that a
jump is safe or technically correct.

\begin{figure}[p]
\centering
\textbf{(a) Input video frame}\\[3pt]
\includegraphics[width=0.90\linewidth]{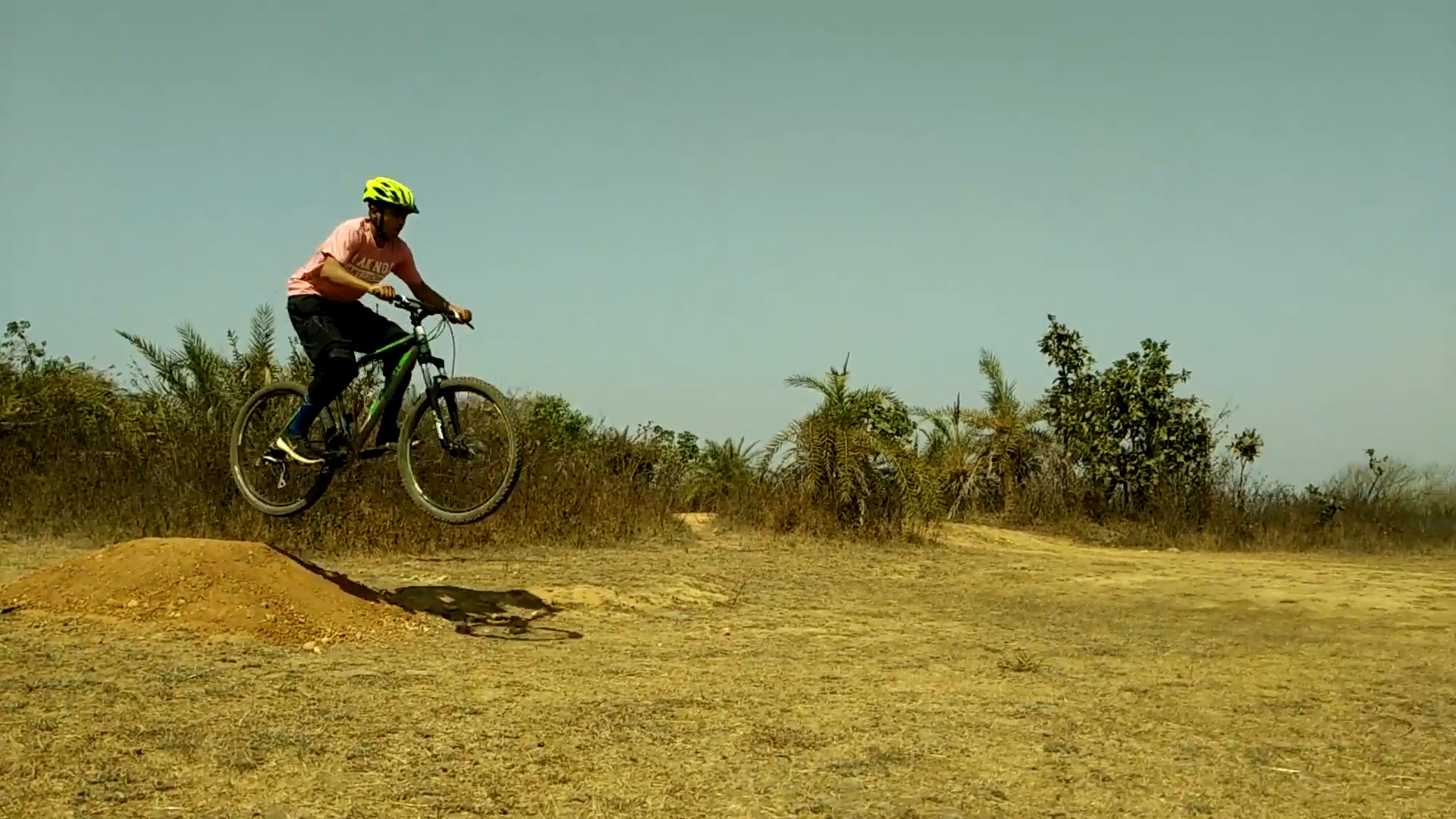}\\[9pt]
\textbf{(b) Recorded system output for the same frame}\\[3pt]
\includegraphics[width=0.90\linewidth]{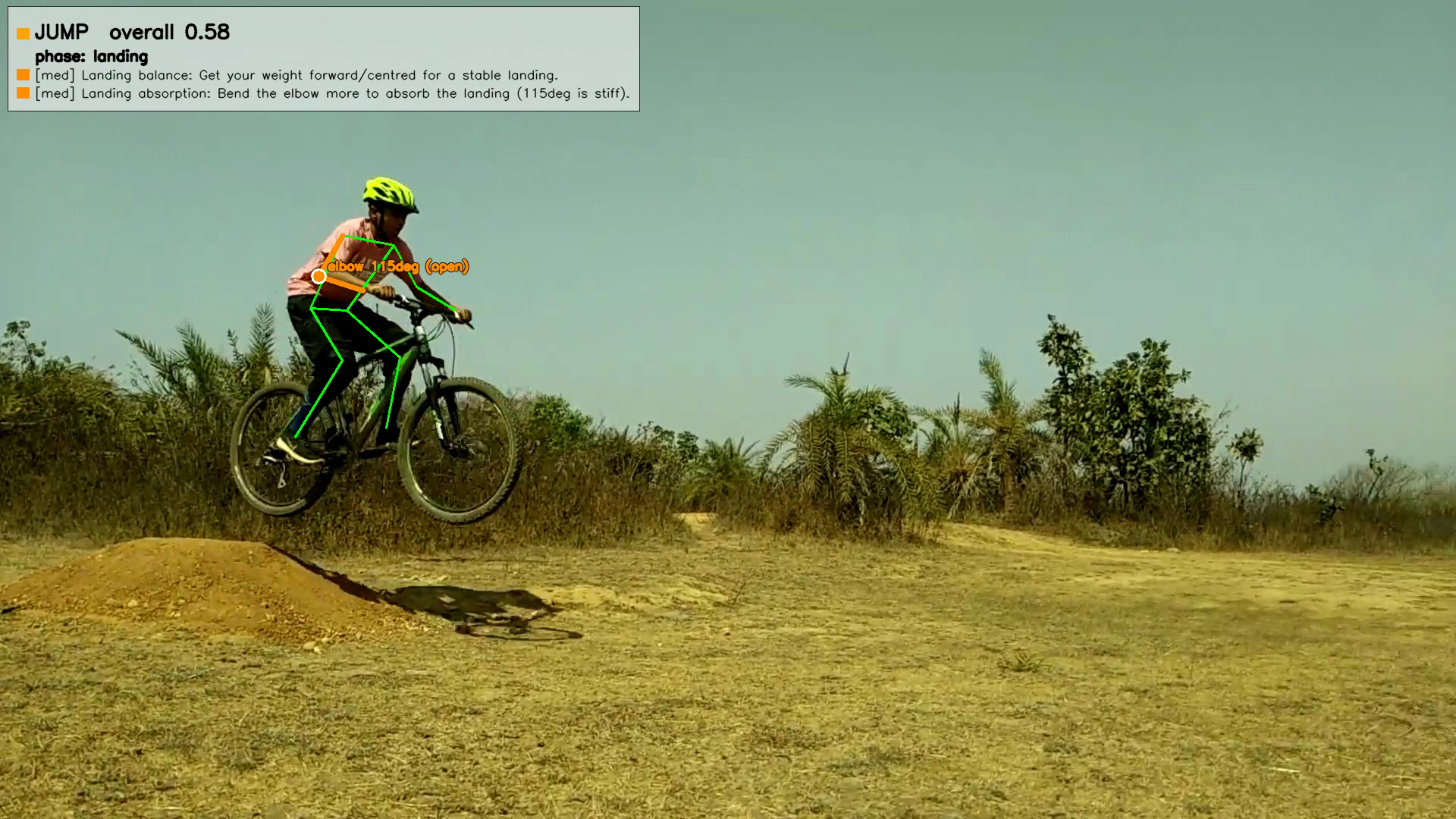}
\caption{Actual full-frame input and recorded output, paired at zero-based
frame 160 (5.33 s at 30 frames/s) of one 181-frame external example.
The green lines are detected pose; the colored elbow marker and text are
automatically generated cues. The displayed phase is ``landing'' although
the rider is visibly airborne. The local annotation says ``open,'' while
the sentence says ``Bend the elbow more.'' These retained inconsistencies
illustrate why the heuristic phase and wording are not ground truth or
validated advice. The 115-degree value is a phase summary from the report,
not a measurement of the single displayed frame. Images are reproduced
with permission.}
\label{fig:recorded-interface}
\end{figure}

Figure~\ref{fig:recorded-interface} gives an actual input/output pair from the
same frame of an external-probe clip. It was extracted from an existing
recorded demonstration; no model was rerun to create this illustration.
The saved report gives overall score 0.579, rounded to 0.58 on the display.
Its eight sub-scores range from 0.577 to 0.580 after three-decimal rounding.
Their near equality is visible evidence of the limited aspect separation in
this example, not eight independently validated assessments. This report
belongs to the earlier demonstration checkpoint, not the canonical
cross-validation model in the main comparison.

\subsection{From a measured deviation to a candidate cue}
Table~\ref{tab:worked-cues} follows four entries in the same saved report.
For the take-off elbow, the measured phase mean is $80.636^\circ$ and the
reference mean is $137.103^\circ$, giving the stored standardized deviation
$z=-2.757$. The selected parameter is therefore below the reference and
the direction field is ``too low.'' The rule selects the elbow and the
take-off phase from the parameter/aspect specification. The emitted
sentence asks the rider to keep the elbow more open. This trace explains
how a cue is assembled; it does not establish that the proposed correction
is appropriate for the rider or the terrain.

\begin{table}[t]
\centering\small
\setlength{\tabcolsep}{5pt}
\begin{tabular}{llrrrr}
\toprule
Aspect / phase & Parameter & Measured & Reference & $z$ & Severity\\
\midrule
Arm position / take-off & Elbow angle ($^\circ$) & 80.636 & 137.103 & $-2.757$ & High\\
Separation / flight & Hip angle ($^\circ$) & 157.122 & 96.104 & 2.556 & High\\
Balance / landing & Fore/aft offset & $-0.0322$ & 0.0004 & $-0.729$ & Medium\\
Absorption / landing & Elbow angle ($^\circ$) & 114.850 & 129.791 & $-0.327$ & Medium\\
\bottomrule
\end{tabular}
\caption{Selected actual cue records for the input/output example in
Figure~\ref{fig:recorded-interface}. Values are phase summaries, rounded
for display; $z$ is the stored deviation before rounding. The fore/aft
offset is dimensionless and camera-relative. Severity is the generator's
rule-based category, not a clinician's or coach's risk rating.}
\label{tab:worked-cues}
\end{table}

The landing-elbow record is an instructive failure. Its absolute deviation
is only 0.327, below the 1.5 deviation threshold, but the aspect score is
about 0.580, below 0.6. The disjunctive trigger can therefore emit this cue
without a large reference deviation. Its direction label and on-screen
wording also disagree about opening or bending the elbow. In addition,
the inferred landing window begins before visible ground contact in the
shown frame. None of these display elements should be read as an
independent annotation of technique.

\subsection{Interpretation of system outputs}
Each cue record contains an aspect, assigned phase, sub-score, severity,
selected joint or body parameter, measured value, reference value,
standardized deviation, direction, and sentence. A downstream interface
should preserve these distinctions. The overall score summarizes the
technique-class training target. The aspect values are unsupervised latent
outputs. The deviation compares image-plane measurements with the
positive-class reference distribution. The sentence is a template-based
candidate recommendation. The external precision and direction results
reported next evaluate some of these outputs; the presence of a visually
plausible overlay is not itself an accuracy result.

\clearpage
\subsection{Cases across the three display bands}
\begin{figure}[p]
\centering
\textbf{Case G: input labelled as good technique}\\[3pt]
\includegraphics[width=0.90\linewidth]{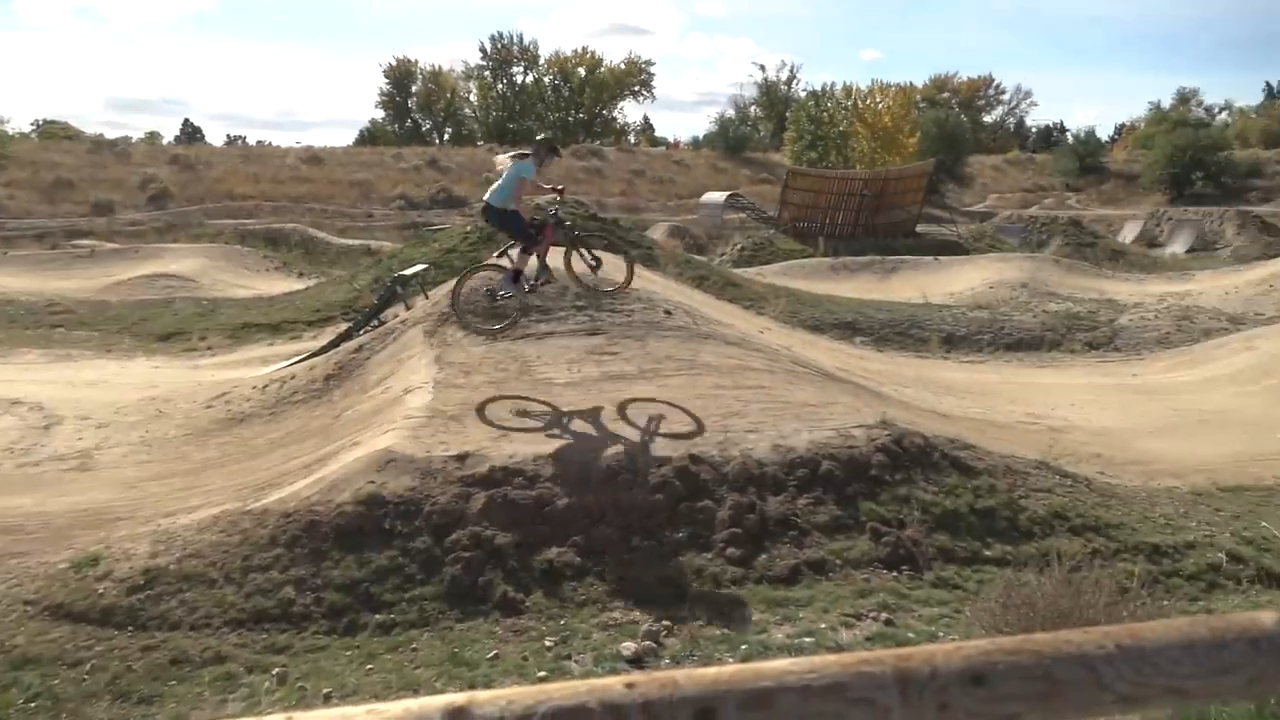}\\[9pt]
\textbf{Case G: recorded green-band output, score 0.875}\\[3pt]
\includegraphics[width=0.90\linewidth]{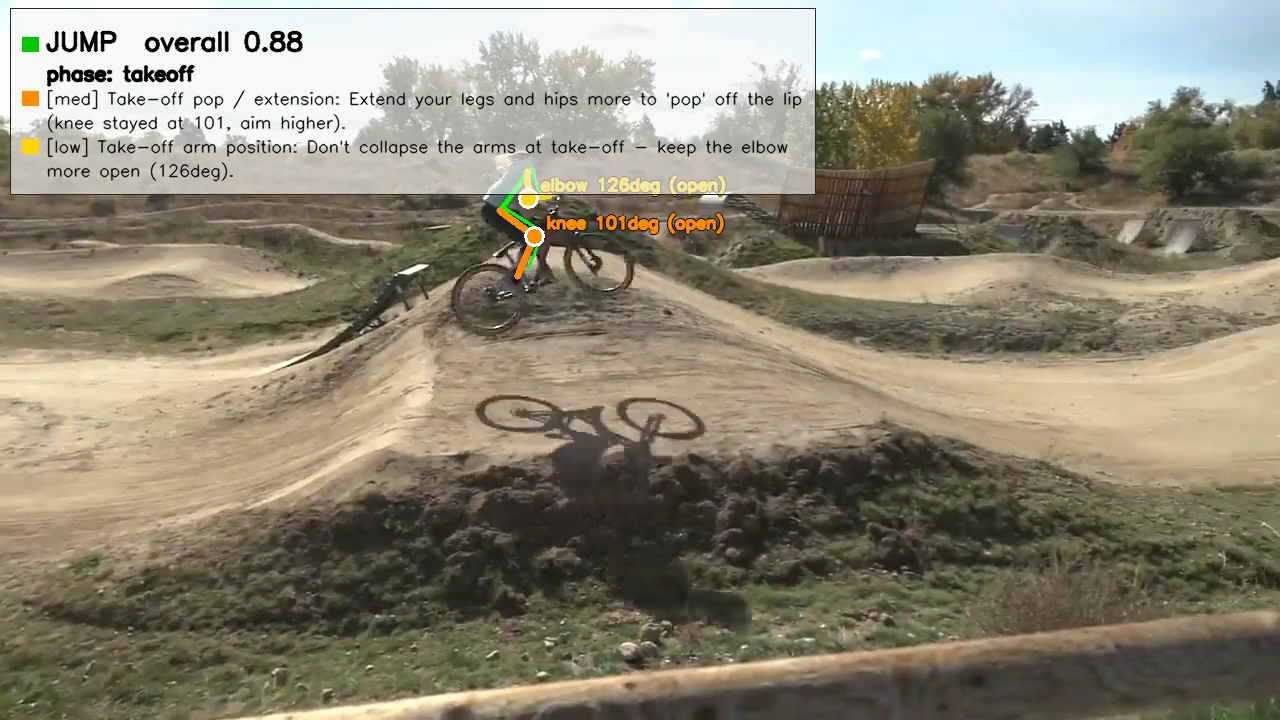}
\caption{High-score display case, paired at frame 24. The overlay rounds
the overall score to 0.88 and shows the assigned take-off phase. The knee
and elbow labels round phase summaries to 101 and 126 degrees; they are
not independent measurements of this single frame. The orange and yellow
cue markers denote medium and low severity, not the green overall band.
The text panel partly covers the rider, an interface limitation preserved
from the recorded output. Images are reproduced with permission.}
\label{fig:green-case}
\end{figure}

\begin{figure}[p]
\centering
\textbf{Case O: input labelled as good technique}\\[3pt]
\includegraphics[width=0.90\linewidth]{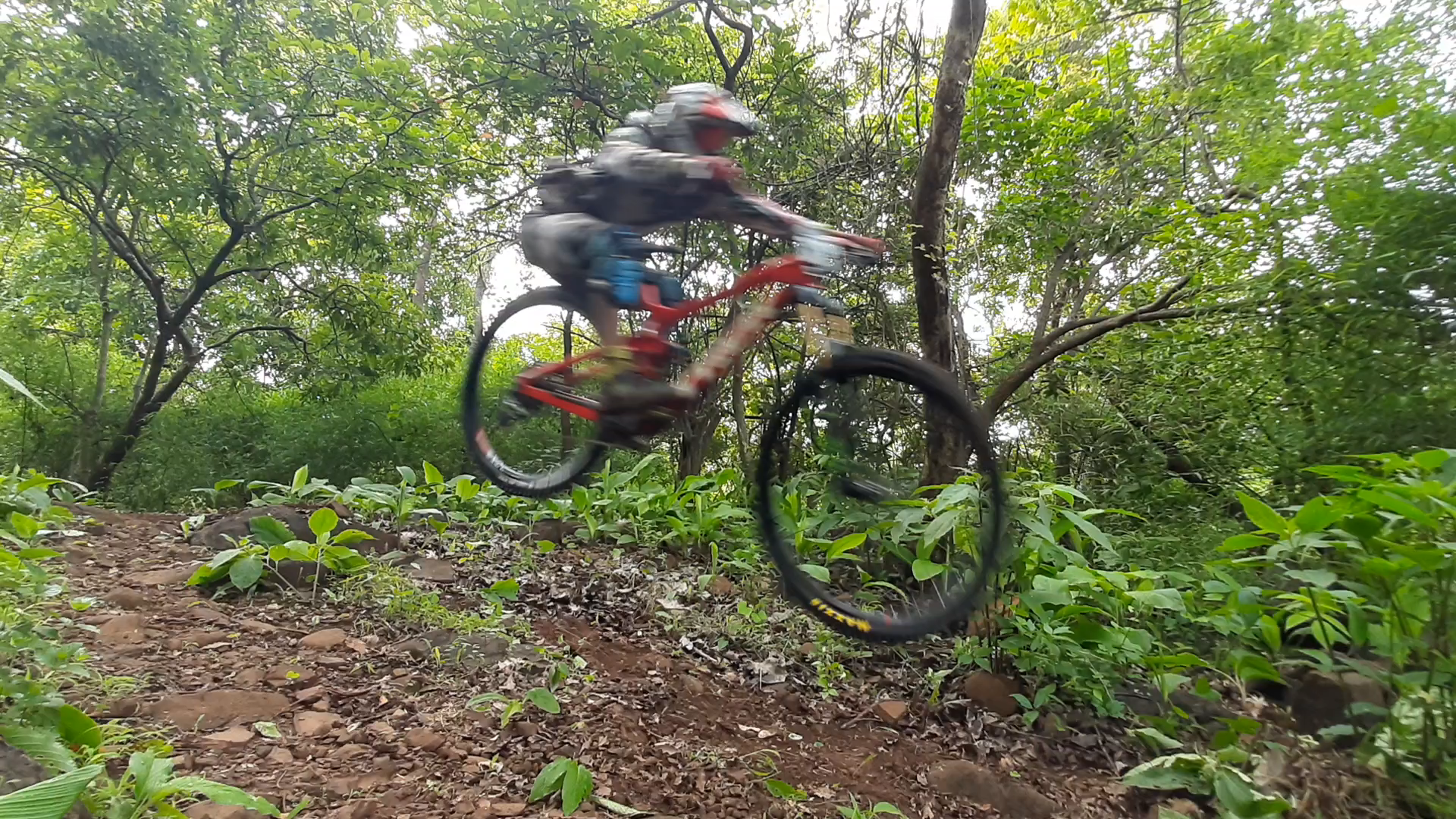}\\[9pt]
\textbf{Case O: recorded orange-band output, score 0.434}\\[3pt]
\includegraphics[width=0.90\linewidth]{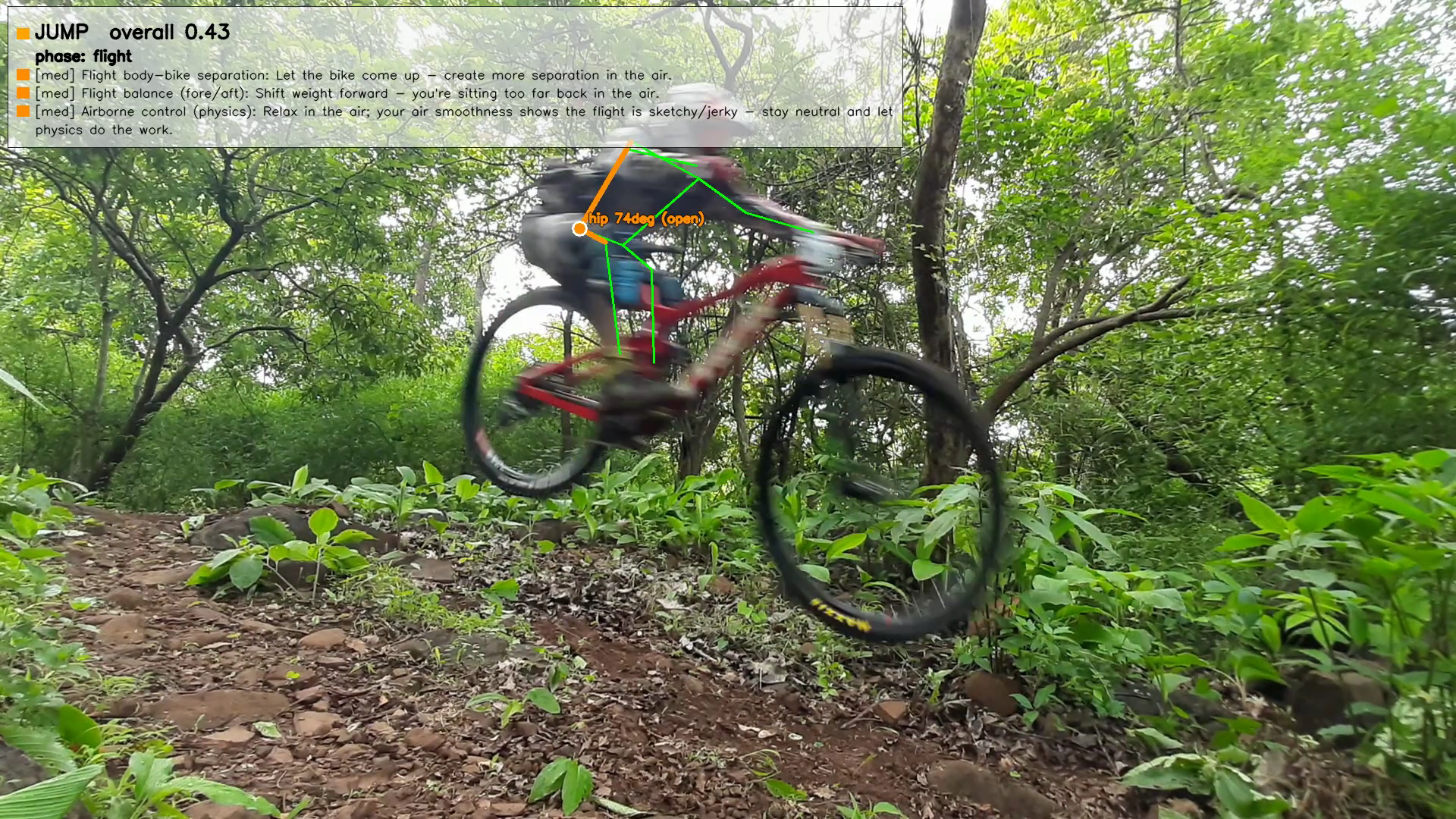}
\caption{Middle-score display case, paired at frame 60. The overall
display rounds to 0.43, while the hip label rounds the flight summary to
74 degrees. The visible motion blur and partially obscured body make the
pose overlay especially important to treat as an estimate. The source
has a good-technique label: orange is an output band, not a medium-quality
ground-truth class. The phase and cue text are generated outputs.
Images are reproduced with permission.}
\label{fig:orange-case}
\end{figure}

\begin{figure}[p]
\centering
\textbf{Case R: input labelled as poor technique}\\[3pt]
\includegraphics[width=0.90\linewidth]{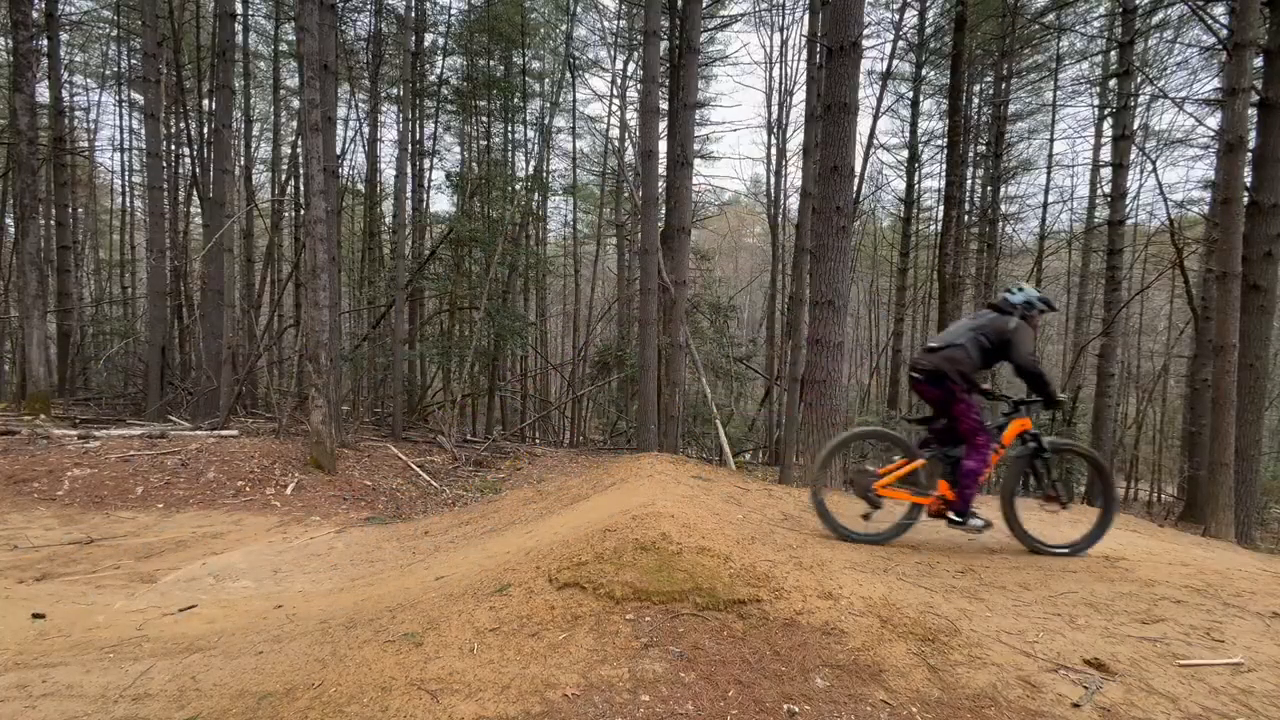}\\[9pt]
\textbf{Case R: recorded red-band output, score 0.256}\\[3pt]
\includegraphics[width=0.90\linewidth]{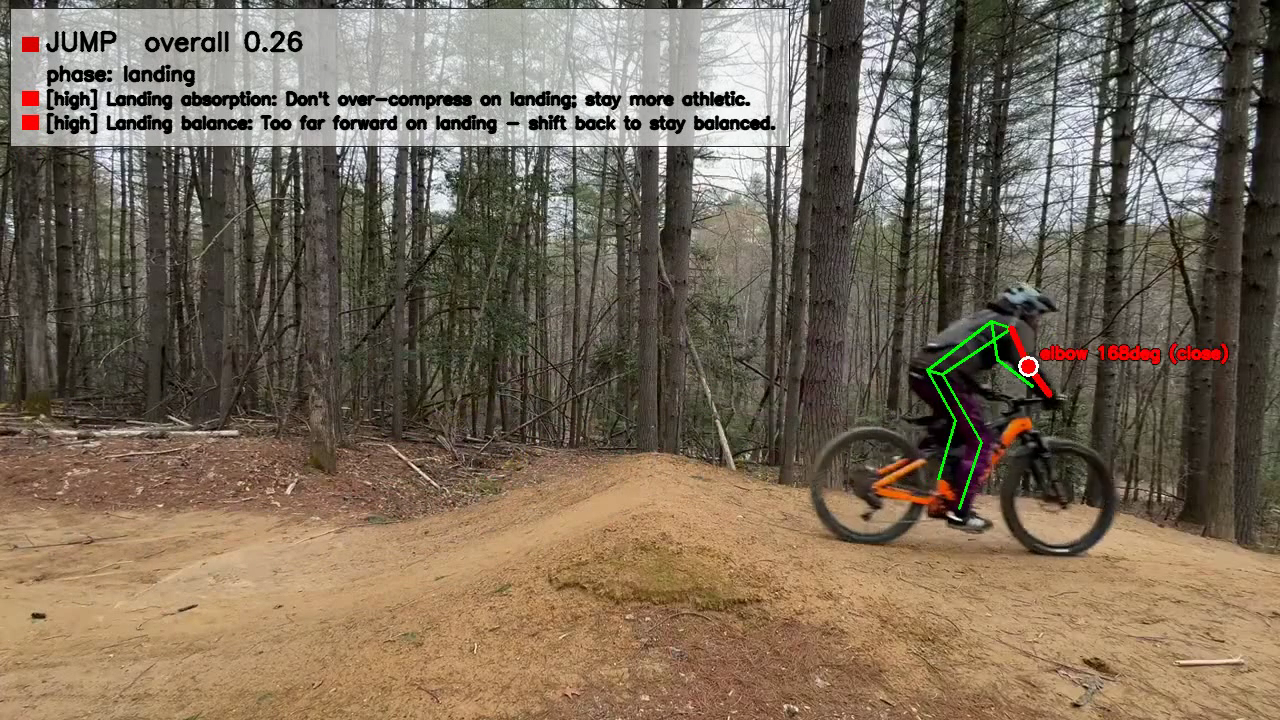}
\caption{Low-score display case, paired at frame 40. The overlay rounds
the overall score to 0.26 and the landing-elbow summary to 168 degrees.
Both displayed landing cues have high severity. The local ``close''
label and the sentence about over-compression are reproduced as generated;
their presence is not evidence of an appropriate corrective action.
Images are reproduced with permission.}
\label{fig:red-case}
\end{figure}

Figures~\ref{fig:green-case}--\ref{fig:red-case} add one recorded input/output
pair for each overall-score display band. The interface uses green for
$s\geq0.6$, orange for $0.4\leq s<0.6$, and red for $s<0.4$.
These bands are display rules, not three annotated quality classes or
calibrated safety levels. They also differ from the binary evaluation
threshold of 0.5. Cue-severity colors have a separate meaning: red is high,
orange is medium, and yellow is low. A green overall banner can therefore
coexist with orange or yellow candidate cues.

The three cases come from existing recorded demonstrations. They were
selected to cover the display bands, not sampled to estimate accuracy.
The green and red clips belong to the main corpus, so their overlays are
not presented as held-out predictions. The orange clip belongs to the
external probe and uses the earlier demonstration scorer. None of these
overlays is attributed to the canonical out-of-fold model in the main
comparison. Original and rendered videos are paired by zero-based frame
index; no model was rerun. The examples preserve the original overlays,
including their limitations.

\begin{table}[ht]
\centering\small
\begin{tabular}{lllrll}
\toprule
Case & Display band & Technique label & Score & Shown phase & Frame\\
\midrule
G & Green & Good technique & 0.875 & Take-off & 24 of 51\\
O & Orange & Good technique & 0.434 & Flight & 60 of 96\\
R & Red & Poor technique & 0.256 & Landing & 40 of 51\\
\bottomrule
\end{tabular}
\caption{The three added demonstration cases. Frame indices are zero-based;
the denominator is the total number of decoded frames. Technique labels are
not independent expert judgments. Case O falls below the binary threshold
despite its good-technique label. These selected examples are not a new test set.}
\label{tab:category-cases}
\end{table}

Table~\ref{tab:category-cases} separates the assigned technique label from the output.
In Case G, the recorded score is 0.875 and all eight rounded aspect scores
are also 0.875. Nevertheless, the system emits eight cues. Its take-off
knee and elbow summaries are $101.050^\circ$ and $126.259^\circ$.
This is a high-scoring display example, not evidence that the aspect
outputs identify independent strengths or that every emitted cue is useful.

Case O has a recorded score of 0.434 and a flight hip summary of
$73.651^\circ$. The rider is visibly motion-blurred in the chosen frame.
The same report contains a landing-elbow summary of only $5.244^\circ$
and a sentence calling it stiff. That combination requires inspection of
the underlying detections and wording; it is not accepted as a validated
biomechanical diagnosis. Case R has a score of 0.256, eight high-severity
cues, and a landing-elbow summary of $168.277^\circ$. Its low score and
poor-technique label agree, but one selected agreement does not establish
classification or corrective-feedback performance.

\clearpage
\clearpage

\clearpage

\clearpage

\subsection{Coverage of phases and feedback categories}
\begin{landscape}
\begin{figure}[p]
\centering
\begin{minipage}{0.46\linewidth}\centering
\includegraphics[width=\linewidth]{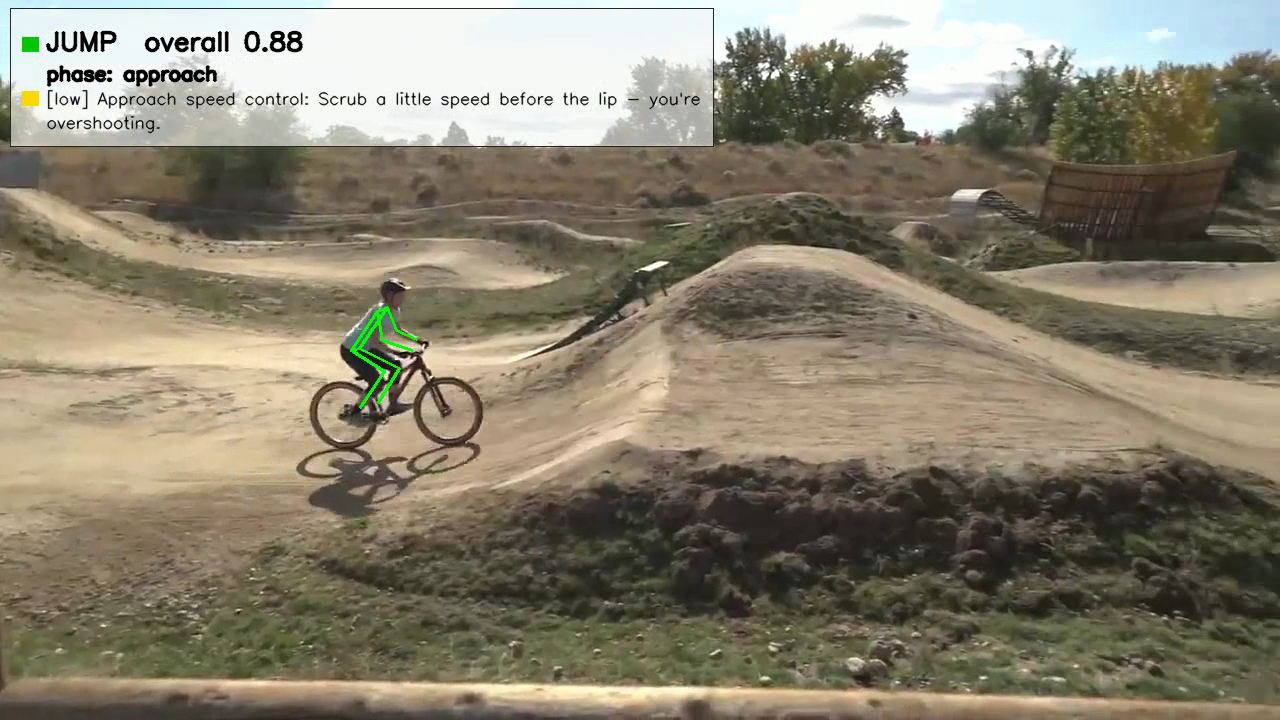}\\
\small (a) Approach display: frame 8
\end{minipage}\hfill
\begin{minipage}{0.46\linewidth}\centering
\includegraphics[width=\linewidth]{assets/green_output24.png}\\
\small (b) Take-off display: frame 24
\end{minipage}\\[8pt]
\begin{minipage}{0.46\linewidth}\centering
\includegraphics[width=\linewidth]{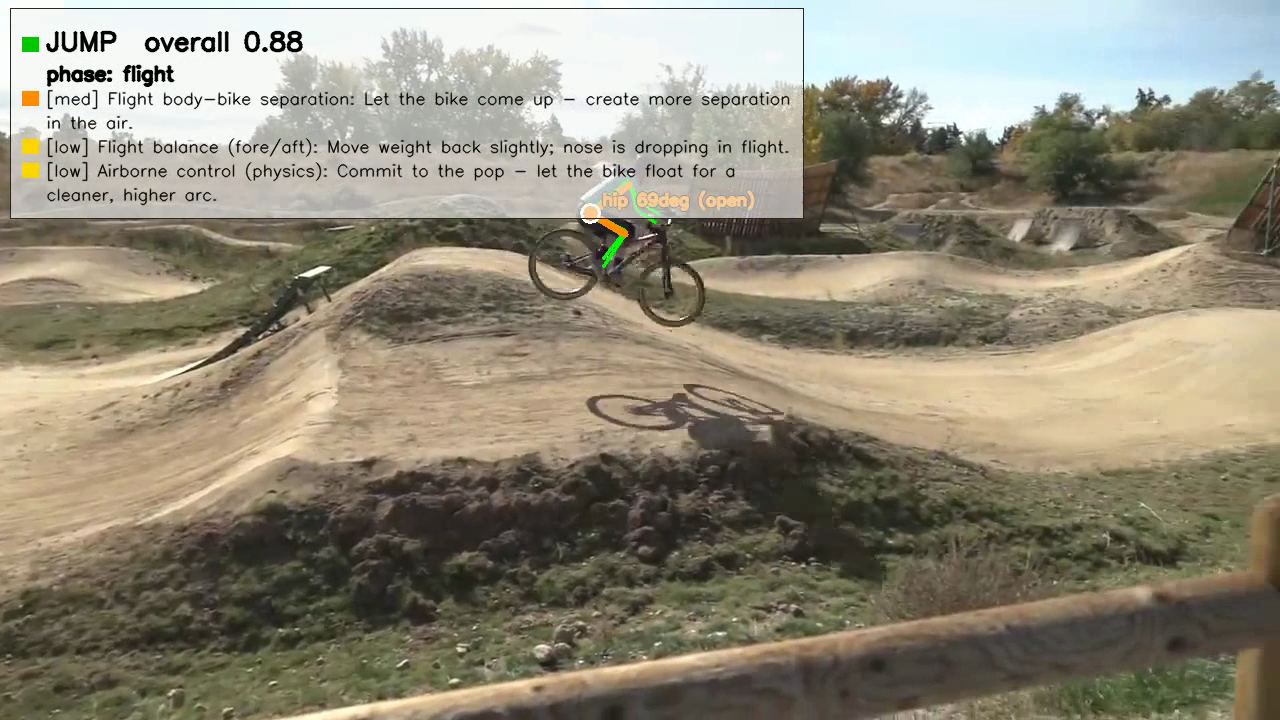}\\
\small (c) Flight display: frame 32
\end{minipage}\hfill
\begin{minipage}{0.46\linewidth}\centering
\includegraphics[width=\linewidth]{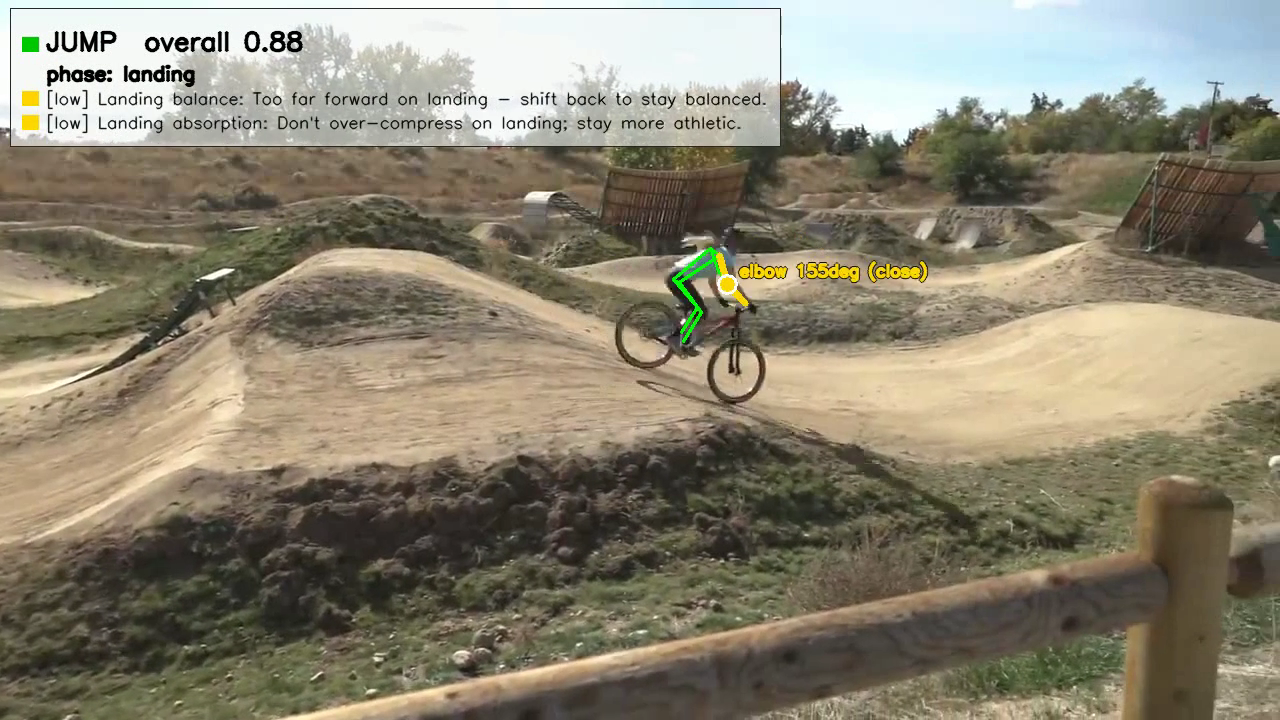}\\
\small (d) Landing display: frame 40
\end{minipage}
\caption{All four recorded phase displays for Case G. The clip-level
score remains 0.88 after display rounding while the visible cue subset
changes with the operational phase. The labels describe the system's
partition, not expert-verified event boundaries. Table~\ref{tab:all-aspect-example}
gives the full set of eight cue measurements. Images are reproduced with permission.}
\label{fig:four-phase-example}
\end{figure}
\end{landscape}

Figure~\ref{fig:four-phase-example} follows Case G through all four
operational phases. Frames 8, 24, 32, and 40 were selected by inspecting
the recorded phase display and keeping the rider visible. This is an
illustrative selection, not independent event annotation.
Table~\ref{tab:all-aspect-example} reports one actual cue for each of the
eight feedback categories from the same saved report. Together, the figure
and table expose every category without treating eight redundant aspect
scores as eight validated assessments.

\begin{table}[ht]
\centering\small
\setlength{\tabcolsep}{4pt}
\renewcommand{\arraystretch}{1.2}
\begin{tabular}{>{\raggedright\arraybackslash}p{0.25\linewidth}>{\raggedright\arraybackslash}p{0.26\linewidth}rrl}
\toprule
Aspect / phase & Selected parameter & Measured & Reference & Severity\\
\midrule
Speed control / approach & Flow-speed proxy & 58.984 & 24.730 & Low\\
Extension / take-off & Knee angle ($^\circ$) & 101.050 & 125.859 & Medium\\
Arm position / take-off & Elbow angle ($^\circ$) & 126.259 & 137.103 & Low\\
Body--bike separation / flight & Hip angle ($^\circ$) & 68.690 & 96.104 & Medium\\
Balance / flight & Torso orientation ($^\circ$) & 54.768 & 44.795 & Low\\
Airborne control / flight & Stored height proxy & 0.000 & 0.031 & Low\\
Absorption / landing & Elbow angle ($^\circ$) & 155.495 & 129.791 & Low\\
Balance / landing & Torso orientation ($^\circ$) & 49.573 & 36.232 & Low\\
\bottomrule
\end{tabular}
\caption{All eight candidate-cue categories for Case G. Values are the
saved phase summaries, rounded to three decimals. Flow speed is an
image-motion proxy in pixels per frame, not rider speed. The height proxy
is a dimensionless image ratio; zero does not establish zero physical
airtime or jump height. Severity is the generator's category, not expert
agreement.}
\label{tab:all-aspect-example}
\end{table}

The corresponding messages ask for less approach speed, more take-off
extension, a more open elbow, more body--bike separation, a backward
weight shift in flight, a stronger pop, less landing compression, and a
backward weight shift on landing. These are summaries of the recorded
candidate text, not recommendations endorsed by this study. Some directions
depend on camera-relative proxies, and their physical interpretation has
not been validated. The low-severity height cue is particularly instructive:
it is emitted even though the stored height proxy is zero and the images
show an airborne rider. This reflects the limitations of the feature's
operational window and the disjunctive cue trigger.

\section{Results}
\label{sec:results}

\subsection{Exploratory in-corpus discrimination}
\label{sec:mainresults}
Table~\ref{tab:models} retains the main binary results. The full scorer, MLP, GRU, and mean-pool transformer used test-fold epoch selection. The adapted CoRe, USDL, and pairwise heads used a fixed 120 epochs; classical estimators did not select an epoch. Feature input also differs by row. The full aspect-query model and GRU/transformer heads receive sequences of video-plus-measurement tokens. Classical models receive mean-pooled clip vectors. Adapted CoRe, USDL, and pairwise ranking receive only mean-pooled V-JEPA 2 vectors. Thus their numbers are not a controlled identical-input architecture comparison. The full aspect-query model scored 0.768 AUROC and 0.812 AP, while adapted pairwise ranking and USDL scored 0.783 and 0.781 AUROC. A mean-pooled transformer scored 0.770. These differences and the protocol do not support superiority of the proposed head. A V-JEPA-only ridge probe scored 0.742 AUROC, whereas a biomechanical-only ridge probe scored 0.487, indicating that frozen video features carry much of the observed binary discrimination in this sample. This does not identify a causal contribution that will persist under rider-disjoint testing.

\begin{table}[t]
\centering
\small
\setlength{\tabcolsep}{4pt}
\begin{tabular}{llrr}
\toprule
Model or feature probe & Input & AUROC & AP \\
\midrule
Ridge regression & Full, pooled & 0.723 & 0.729 \\
RBF support-vector regression & Full, pooled & 0.740 & 0.733 \\
Random forest & Full, pooled & 0.727 & 0.749 \\
Gradient boosting & Full, pooled & 0.597 & 0.637 \\
V-JEPA 2 features, ridge & Video, pooled & 0.742 & 0.758 \\
Biomechanical features, ridge & Measured, pooled & 0.487 & 0.551 \\
MLP, mean pooling & Full sequence & 0.729 & 0.744 \\
GRU & Full sequence & 0.762 & 0.805 \\
Transformer, mean pooling & Full sequence & 0.770 & 0.786 \\
Adapted CoRe~\cite{yu2021core} & Video, pooled & 0.763 & 0.769 \\
Adapted USDL~\cite{tang2020uncertainty} & Video, pooled & 0.781 & 0.811 \\
Adapted pairwise ranking~\cite{doughty2018whosbetter} & Video, pooled & 0.783 & 0.808 \\
Full aspect-query scorer & Full sequence & 0.768 & 0.812 \\
\bottomrule
\end{tabular}
\caption{Pooled out-of-fold binary metrics on 95 clips, one seed, five clip-level folds. ``Full'' means video and measured features; ``video'' means frozen V-JEPA 2 features; ``pooled'' means an average over four phases. The full, MLP, GRU, and transformer heads use the held-out fold for epoch selection; adapted CoRe, USDL, and pairwise heads fit for 120 fixed epochs. Classical estimators have no epoch selection. Adapted prior methods use binary technique labels, not their original expert-score tasks.}
\label{tab:models}
\end{table}

At score threshold 0.5, the full scorer classified 34 of 52 clips labelled as good technique and 30 of 43 clips labelled as poor technique correctly in the pooled out-of-fold predictions (Table~\ref{tab:confusion}). The resulting accuracy is $64/95=0.674$, with positive-class precision $34/47=0.723$ and recall $34/52=0.654$. Across the five test-selected folds, mean accuracy was $0.674\pm0.117$ and mean Spearman association with the binary label was $0.545\pm0.206$. These describe the same limited in-corpus protocol; the high variance across folds and the label ties make them unsuitable as continuous-quality evidence.

\begin{table}[t]
\centering
\small
\begin{tabular}{lrr}
\toprule
 & Predicted good technique & Predicted poor technique \\
\midrule
Labelled good technique & 34 & 18 \\
Labelled poor technique & 13 & 30 \\
\bottomrule
\end{tabular}
\caption{Full scorer's pooled out-of-fold decisions at threshold 0.5. Epochs were selected using their reporting folds; the entries are not an independent test-set confusion matrix.}
\label{tab:confusion}
\end{table}

The original fold-specific mean predictor produced a different training-fold prevalence in each fold. Its pooled AUROC was 0.475 and AP 0.535. That AUROC is an artifact of ordering the five prevalence constants after pooling, rather than meaningful discrimination by a constant predictor. A single constant score has AUROC 0.5 by convention. We do not use the pooled value to establish a performance margin over a floor.

\subsection{Input and head ablations}
Table~\ref{tab:ablations} shows the ablations under the same exploratory protocol. Dropping biomechanical scoring inputs raised AUROC from 0.768 to 0.809, though AP fell slightly from 0.812 to 0.803. This is evidence against a claim that the biomechanical inputs improved binary scoring in this run. It does not test the separate reference comparison for cues. Removing the frozen video features reduced AUROC to 0.619. Mean pooling without aspect queries yielded 0.770. The ``phase-pooled'' variant replaces phase inputs with a repeated pooled token while retaining learned positional embeddings; its 0.763 AUROC cannot be interpreted as an ablation of all phase information.

\begin{table}[t]
\centering
\small
\begin{tabular}{lrr}
\toprule
Variant & AUROC & AP \\
\midrule
Full aspect-query scorer & 0.768 & 0.812 \\
No frozen video features & 0.619 & 0.658 \\
No biomechanical scoring inputs & 0.809 & 0.803 \\
No flight descriptors & 0.734 & 0.789 \\
Phase-pooled inputs & 0.763 & 0.745 \\
No pairwise ranking term & 0.753 & 0.749 \\
Mean pooling, no aspect queries & 0.770 & 0.786 \\
\bottomrule
\end{tabular}
\caption{Single-seed ablations with the same clip-level folds and test-epoch-selected reporting protocol as Table~\ref{tab:models}. Values are pooled out-of-fold binary metrics.}
\label{tab:ablations}
\end{table}

\subsection{Selection and aspect diagnostics}
The separate epoch-selection diagnostic obtained AUROC/AP of 0.781/0.826 when the test fold selected the epoch, and 0.678/0.699 when an inner split selected it. The first value is from a separate run and is not the 0.768 full-model result in Table~\ref{tab:models}. The 0.103 AUROC difference mixes test-fold selection with an approximate reduction from 76 to 57 training clips per outer fold. We cannot assign a portion to selection optimism or use 0.678 as a matched lower bound. The experiment instead shows that the reported model is sensitive to the evaluation and training protocol. A valid estimate requires inner selection followed by fresh training on the complete outer-training partition, with the outer test fold untouched.

The eight unsupervised aspect queries did not yield distinct interpretations. Across three seeds, the average normalized entropy of mixture weights was approximately 1.00 and the largest weight approximately 0.126, close to the uniform value 0.125. Mean absolute correlation among aspect sub-scores was approximately 1.00. The mean top-one agreement between query attention and each aspect's designated phase was 0.25, equal to four-way chance. Thus attention and sub-scores cannot be used as evidence for aspect-level explanation or temporal localization in this corpus.

\subsection{External scoring and candidate feedback}
\label{sec:external}
The earlier checkpoint's 17-clip external scoring probe gave binary accuracy 0.412 against technique labels at a 0.5 score threshold and Spearman association 0.132 against those labels. Against the 17 expert scores, Spearman association was $-0.135$ and mean absolute error was 0.309 after dividing expert scores by ten to place them on the unit interval. The small sample and possible distribution shift limit precision, but the observed result gives no support for external scoring utility. Because this checkpoint differs from the one in Tables~\ref{tab:models} and~\ref{tab:ablations}, those in-corpus and external numbers must not be read as a before/after performance comparison of one model.

The cue generator emitted 136 cues across the 17 clips, or eight per clip on average. Relative to each coach, 16 cues matched and 120 were unmatched, with no missed annotated cues under the clip--aspect matching rule. Each coach therefore yielded precision 0.118, recall 1.000, and $F_1$ 0.211. Among the 16 matched cues for each coach, joint accuracy was 0.375. Direction accuracy was 0.688 for coach A and 0.375 for coach B; phase agreement was 1.000 and 0.938, respectively. The phase is assigned from the aspect specification, so these phase figures do not validate learned temporal localization. Across 136 assessed items, coach agreement was 0.853 by raw proportion but Cohen's $\kappa$ was 0.292. The direction result is annotator-dependent, and the low precision and joint accuracy make these cues exploratory candidates rather than validated advice.

An optional InternVL3-based cue rewording and direction step~\cite{zhu2025internvlthree} was tested on 20 cues. It changed eight directions (40\%), with no fallback in that small sample and mean latency 4.44 seconds per cue. No adjudicated outcome establishes whether the changes improved direction accuracy. This optional step was not included in the primary cue figures above.

\subsection{Training-set-size probe}
Table~\ref{tab:efficiency} summarizes an additional three-seed subsampling experiment. Stratified subsets of 20, 40, 60, 80, and 95 clips were drawn with subsample seeds 0, 1, and 2; each subset received its own five clip-level folds. Neural training used seeds 42, 43, and 44 and a maximum of 60 epochs, unlike the 120-epoch main comparison. It reports Spearman association with the binary technique target, not continuous quality. At 95 clips, mean values were 0.439 for the aspect-query model, 0.444 for mean-pool transformer, 0.468 for adapted CoRe, and 0.417 for frozen-feature ridge. The curves are nonmonotone and vary across three draws. They are descriptive only; no paired uncertainty analysis establishes equivalence, an advantage at a given sample size, or a plateau.

\begin{table}[t]
\centering
\small
\setlength{\tabcolsep}{5pt}
\begin{tabular}{rcccc}
\toprule
$N$ & Aspect query & Transformer & CoRe & Video ridge \\
\midrule
20 & $0.206\pm0.143$ & $0.183\pm0.213$ & $0.369\pm0.259$ & $0.381\pm0.228$ \\
40 & $0.360\pm0.035$ & $0.311\pm0.010$ & $0.280\pm0.025$ & $0.324\pm0.047$ \\
60 & $0.329\pm0.108$ & $0.317\pm0.172$ & $0.421\pm0.214$ & $0.446\pm0.196$ \\
80 & $0.434\pm0.044$ & $0.427\pm0.028$ & $0.448\pm0.053$ & $0.402\pm0.038$ \\
95 & $0.439\pm0.016$ & $0.444\pm0.011$ & $0.468\pm0.023$ & $0.417\pm0.000$ \\
\bottomrule
\end{tabular}
\caption{Mean $\pm$ standard deviation of association with binary technique labels across three subsampling seeds. The probe uses a separate training schedule and is not numerically identical to Table~\ref{tab:models}. No method equivalence or saturation is inferred.}
\label{tab:efficiency}
\end{table}

The scorer head contains 485,898 trainable parameters. End-to-end latency, including video decoding, pose and bike detection, phase segmentation, and frozen video feature extraction, was not established.

\section{Discussion, Limitations, and Future Work}
\label{sec:discussion}
The in-corpus values show that frozen video features can separate these particular technique labels to some extent under the reported exploratory protocol. They do not establish that the scorer measures jump quality. The labels contain only two values, several heads including the proposed scorer used the test fold to choose an epoch, and the adapted prior methods received different inputs from the full scorer. The two arms of the subsequent diagnostic had unequal training sizes. A comparative conclusion would require a consistent model-selection protocol and an explicit control of input features for all models.

The independence unit is also unresolved. Rider, session, trail, and camera identities are missing, and repeated appearance or capture conditions may carry label information. The present clip-level folds cannot quantify new-rider or new-location performance. Future collection should record those identities, identify near-duplicate clips, and reserve rider- or session-disjoint test sets. A fixed, predeclared epoch-selection procedure should be run on every model, with retraining on the complete outer-training set after selection. Fold-level or calibrated score aggregation and uncertainty over riders or sessions would then be more meaningful than the pooled clip-level comparisons here.

The feedback probe does not establish corrective validity. Image-plane posture depends on camera view; phase boundaries are heuristic; the reference distribution is derived from technique labels rather than adjudicated technique; and each cue is assembled by a rule that can trigger frequently. The deficiency-output head has no cue-level supervision, which further limits the meaning of its trigger. The phase agreement figure partly reflects the built-in aspect-to-phase assignment. The two coaches disagreed materially about direction, and only 16 matched cues per coach were available for conditional direction and joint checks. The separate external scorer had poor measured agreement. A subsequent study should obtain independent phase boundaries, joint and direction labels, and continuous quality judgments from multiple coaches, then adjudicate disagreement before testing whether a cue improves rider performance. Threshold calibration should target useful precision as well as coverage. Any deployment would also require an end-to-end timing and failure analysis that includes perception.

Finally, the eight aspect scores are redundant without aspect supervision. The current evidence provides no support for reading their weights or attention as explanations. Future work can test explicit aspect annotations, simpler single-score heads, and whether a reference comparison adds value after pose accuracy and viewpoint effects are independently measured. The results here set out a limited case study: an exploratory binary signal within one small corpus and a candidate-cue mechanism whose external validity remains open.

\bibliographystyle{plain}
\bibliography{main}

@inproceedings{parmar2019whatandhow,
  author = {Parmar, Paritosh and Morris, Brendan Tran},
  title = {What and How Well You Performed? A Multitask Learning Approach to Action Quality Assessment},
  booktitle = {Proceedings of the IEEE/CVF Conference on Computer Vision and Pattern Recognition},
  year = {2019}
}

@inproceedings{tang2020uncertainty,
  author = {Tang, Yansong and Ni, Zanlin and Zhou, Jiahuan and Zhang, Danyang and Lu, Jiwen and Wu, Ying and Zhou, Jie},
  title = {Uncertainty-Aware Score Distribution Learning for Action Quality Assessment},
  booktitle = {Proceedings of the IEEE/CVF Conference on Computer Vision and Pattern Recognition},
  year = {2020}
}

@inproceedings{yu2021core,
  author = {Yu, Xumin and Rao, Yongming and Zhao, Wenliang and Lu, Jiwen and Zhou, Jie},
  title = {Group-Aware Contrastive Regression for Action Quality Assessment},
  booktitle = {Proceedings of the IEEE/CVF International Conference on Computer Vision},
  year = {2021}
}

@inproceedings{doughty2018whosbetter,
  author = {Doughty, Hazel and Damen, Dima and Mayol-Cuevas, Walterio},
  title = {Who's Better? Who's Best? Pairwise Deep Ranking for Skill Determination},
  booktitle = {Proceedings of the IEEE/CVF Conference on Computer Vision and Pattern Recognition},
  year = {2018}
}

@inproceedings{xu2022finediving,
  author = {Xu, Jinglin and Rao, Yongming and Yu, Xumin and Chen, Guangyi and Zhou, Jie and Lu, Jiwen},
  title = {{FineDiving}: A Fine-Grained Dataset for Procedure-Aware Action Quality Assessment},
  booktitle = {Proceedings of the IEEE/CVF Conference on Computer Vision and Pattern Recognition},
  year = {2022}
}

@inproceedings{dadashzadeh2024pecop,
  author = {Dadashzadeh, Amirhossein and Duan, Shuchao and Whone, Alan and Mirmehdi, Majid},
  title = {{PECoP}: Parameter Efficient Continual Pretraining for Action Quality Assessment},
  booktitle = {Proceedings of the IEEE/CVF Winter Conference on Applications of Computer Vision},
  year = {2024},
  pages = {42--52},
  doi = {10.1109/WACV57701.2024.00012},
  url = {https://openaccess.thecvf.com/content/WACV2024/html/Dadashzadeh_PECoP_Parameter_Efficient_Continual_Pretraining_for_Action_Quality_Assessment_WACV_2024_paper.html}
}

@misc{matsuyama2023iris,
  author = {Matsuyama, Hitoshi and Kawaguchi, Nobuo and Lim, Brian Y.},
  title = {{IRIS}: Interpretable Rubric-Informed Segmentation for Action Quality Assessment},
  year = {2023},
  howpublished = {arXiv:2303.09097},
  note = {28th International Conference on Intelligent User Interfaces},
  doi = {10.48550/arXiv.2303.09097}
}

@misc{singh2024poze,
  author = {Singh, Agamdeep and {Sujit PB} and Vatsa, Mayank},
  title = {{Poze}: Sports Technique Feedback under Data Constraints},
  year = {2024},
  howpublished = {arXiv:2411.05734},
  doi = {10.48550/arXiv.2411.05734}
}

@inproceedings{ashutosh2025expertaf,
  author = {Ashutosh, Kumar and Nagarajan, Tushar and Pavlakos, Georgios and Kitani, Kris and Grauman, Kristen},
  title = {{ExpertAF}: Expert Actionable Feedback from Video},
  booktitle = {Proceedings of the IEEE/CVF Conference on Computer Vision and Pattern Recognition},
  year = {2025},
  pages = {13582--13594},
  doi = {10.1109/CVPR52734.2025.01268},
  url = {https://openaccess.thecvf.com/content/CVPR2025/html/Ashutosh_ExpertAF_Expert_Actionable_Feedback_from_Video_CVPR_2025_paper.html}
}

@inproceedings{okamoto2024hierarchical,
  author = {Okamoto, Lauren and Parmar, Paritosh},
  title = {Hierarchical NeuroSymbolic Approach for Comprehensive and Explainable Action Quality Assessment},
  booktitle = {Proceedings of the IEEE/CVF Conference on Computer Vision and Pattern Recognition Workshops},
  year = {2024}
}

@misc{assran2025vjepa2,
  author = {Assran, Mido and Bardes, Adrien and Fan, David and Garrido, Quentin and Howes, Russell and Komeili, Mojtaba and Muckley, Matthew and Rizvi, Ammar and Roberts, Claire and Sinha, Koustuv and Zholus, Artem and others},
  title = {{V-JEPA 2}: Self-Supervised Video Models Enable Understanding, Prediction and Planning},
  year = {2025},
  howpublished = {arXiv:2506.09985},
  doi = {10.48550/arXiv.2506.09985}
}

@misc{zhu2025internvlthree,
  author = {Zhu, Jinguo and Wang, Weiyun and Chen, Zhe and Liu, Zhaoyang and Ye, Shenglong and Gu, Lixin and Duan, Yuchen and Tian, Hao and Su, Weijie and Shao, Jie and others},
  title = {{InternVL3}: Exploring Advanced Training and Test-Time Recipes for Open-Source Multimodal Models},
  year = {2025},
  howpublished = {arXiv:2504.10479},
  doi = {10.48550/arXiv.2504.10479}
}

@inproceedings{parmar2019aqa7,
  author = {Parmar, Paritosh and Morris, Brendan Tran},
  title = {Action Quality Assessment Across Multiple Actions},
  booktitle = {IEEE Winter Conference on Applications of Computer Vision},
  year = {2019},
  url = {https://arxiv.org/abs/1812.06367}
}

@inproceedings{xu2022gdlt,
  author = {Xu, Angchi and Zeng, Ling-An and Zheng, Wei-Shi},
  title = {Likert Scoring With Grade Decoupling for Long-Term Action Assessment},
  booktitle = {Proceedings of the IEEE/CVF Conference on Computer Vision and Pattern Recognition},
  year = {2022},
  pages = {3232--3241}
}

@inproceedings{wang2021tsa,
  author = {Wang, Shunli and Yang, Dingkang and Zhai, Peng and Chen, Chixiao and Zhang, Lihua},
  title = {{TSA-Net}: Tube Self-Attention Network for Action Quality Assessment},
  booktitle = {Proceedings of the 29th ACM International Conference on Multimedia},
  year = {2021},
  doi = {10.1145/3474085.3475438}
}

@inproceedings{li2022pcln,
  author = {Li, Mingzhe and Zhang, Hong-Bo and Lei, Qing and Fan, Zongwen and Liu, Jinghua and Du, Ji-Xiang},
  title = {Pairwise Contrastive Learning Network for Action Quality Assessment},
  booktitle = {Computer Vision---ECCV 2022},
  year = {2022},
  pages = {457--473},
  url = {https://www.ecva.net/papers/eccv_2022/papers_ECCV/html/6731_ECCV_2022_paper.php}
}

@inproceedings{zhang2024nae,
  author = {Zhang, Shiyi and Bai, Sule and Chen, Guangyi and Chen, Lei and Lu, Jiwen and Wang, Junle and Tang, Yansong},
  title = {Narrative Action Evaluation with Prompt-Guided Multimodal Interaction},
  booktitle = {Proceedings of the IEEE/CVF Conference on Computer Vision and Pattern Recognition},
  year = {2024},
  pages = {18430--18439}
}

@inproceedings{han2025finecausal,
  author = {Han, Ruisheng and Zhou, Kanglei and Atapour-Abarghouei, Amir and Liang, Xiaohui and Shum, Hubert P. H.},
  title = {{FineCausal}: A Causal-Based Framework for Interpretable Fine-Grained Action Quality Assessment},
  booktitle = {Proceedings of the IEEE/CVF Conference on Computer Vision and Pattern Recognition Workshops},
  year = {2025},
  pages = {6064--6073}
}

@inproceedings{grauman2024egoexo4d,
  author = {Grauman, Kristen and Westbury, Andrew and Torresani, Lorenzo and Kitani, Kris and Malik, Jitendra and Afouras, Triantafyllos and Ashutosh, Kumar and others},
  title = {{Ego-Exo4D}: Understanding Skilled Human Activity from First- and Third-Person Perspectives},
  booktitle = {Proceedings of the IEEE/CVF Conference on Computer Vision and Pattern Recognition},
  year = {2024},
  pages = {19383--19400}
}

@inproceedings{tong2022videomae,
  author = {Tong, Zhan and Song, Yibing and Wang, Jue and Wang, Limin},
  title = {{VideoMAE}: Masked Autoencoders are Data-Efficient Learners for Self-Supervised Video Pre-Training},
  booktitle = {Advances in Neural Information Processing Systems},
  volume = {35},
  year = {2022}
}

@misc{wang2022internvideo,
  author = {Wang, Yi and Li, Kunchang and Li, Yizhuo and He, Yinan and Huang, Bingkun and Zhao, Zhiyu and Zhang, Hongjie and Xu, Jilan and Liu, Yi and Wang, Zun and Xing, Sen and Chen, Guo and Pan, Junting and Yu, Jiashuo and Wang, Yali and Wang, Limin and Qiao, Yu},
  title = {{InternVideo}: General Video Foundation Models via Generative and Discriminative Learning},
  howpublished = {arXiv:2212.03191},
  year = {2022},
  doi = {10.48550/arXiv.2212.03191}
}

@misc{bardes2024vjepa,
  author = {Bardes, Adrien and Garrido, Quentin and Ponce, Jean and Chen, Xinlei and Rabbat, Michael and LeCun, Yann and Assran, Mahmoud and Ballas, Nicolas},
  title = {Revisiting Feature Prediction for Learning Visual Representations from Video},
  howpublished = {arXiv:2404.08471},
  year = {2024},
  doi = {10.48550/arXiv.2404.08471}
}

@inproceedings{liu2024llava,
  author = {Liu, Haotian and Li, Chunyuan and Wu, Qingyang and Lee, Yong Jae},
  title = {Visual Instruction Tuning},
  booktitle = {Advances in Neural Information Processing Systems},
  volume = {36},
  year = {2023}
}

@misc{chen2024internvl,
  author = {Chen, Zhe and Wu, Jiannan and Wang, Wenhai and Su, Weijie and Chen, Guo and Xing, Sen and Zhong, Muyan and Zhang, Qinglong and Zhu, Xizhou and Lu, Lewei and Li, Bin and Luo, Ping and Lu, Tong and Qiao, Yu and Dai, Jifeng},
  title = {{InternVL}: Scaling up Vision Foundation Models and Aligning for Generic Visual-Linguistic Tasks},
  howpublished = {arXiv:2312.14238},
  year = {2023},
  note = {Revised January 2024},
  doi = {10.48550/arXiv.2312.14238}
}

@misc{gemmateam2024gemma,
  author = {{Gemma Team}},
  title = {{Gemma}: Open Models Based on Gemini Research and Technology},
  howpublished = {arXiv:2403.08295},
  year = {2024},
  doi = {10.48550/arXiv.2403.08295}
}

@inproceedings{maji2022yolopose,
  author = {Maji, Debapriya and Nagori, Soyeb and Mathew, Manu and Poddar, Deepak},
  title = {{YOLO-Pose}: Enhancing {YOLO} for Multi Person Pose Estimation Using Object Keypoint Similarity Loss},
  booktitle = {Proceedings of the IEEE/CVF Conference on Computer Vision and Pattern Recognition Workshops},
  year = {2022},
  pages = {2637--2646}
}

@inproceedings{pavllo2019videopose3d,
  author = {Pavllo, Dario and Feichtenhofer, Christoph and Grangier, David and Auli, Michael},
  title = {{3D} Human Pose Estimation in Video With Temporal Convolutions and Semi-Supervised Training},
  booktitle = {Proceedings of the IEEE/CVF Conference on Computer Vision and Pattern Recognition},
  year = {2019},
  pages = {7753--7762}
}

@inproceedings{yan2018stgcn,
  author = {Yan, Sijie and Xiong, Yuanjun and Lin, Dahua},
  title = {Spatial Temporal Graph Convolutional Networks for Skeleton-Based Action Recognition},
  booktitle = {Proceedings of the AAAI Conference on Artificial Intelligence},
  volume = {32},
  year = {2018},
  doi = {10.1609/aaai.v32i1.12328}
}

@inproceedings{zhu2023motionbert,
  author = {Zhu, Wentao and Ma, Xiaoxuan and Liu, Zhaoyang and Liu, Libin and Wu, Wayne and Wang, Yizhou},
  title = {{MotionBERT}: A Unified Perspective on Learning Human Motion Representations},
  booktitle = {Proceedings of the IEEE/CVF International Conference on Computer Vision},
  year = {2023},
  pages = {15085--15099}
}

@misc{leuthold2025physics,
  author = {Leuthold, Tobias and Xiloyannis, Michele and Zimmermann, Yves},
  title = {Physics Informed Human Posture Estimation Based on {3D} Landmarks from Monocular {RGB}-Videos},
  howpublished = {arXiv:2512.06783},
  year = {2025},
  doi = {10.48550/arXiv.2512.06783}
}

@article{uhlrich2023opencap,
  author = {Uhlrich, Scott D. and Falisse, Antoine and Kidzi{\'n}ski, {\L}ukasz and Muccini, Julie and Ko, Michael and Chaudhari, Akshay S. and Hicks, Jennifer L. and Delp, Scott L.},
  title = {{OpenCap}: Human Movement Dynamics from Smartphone Videos},
  journal = {PLOS Computational Biology},
  volume = {19},
  number = {10},
  pages = {e1011462},
  year = {2023},
  doi = {10.1371/journal.pcbi.1011462}
}

@misc{kakavand2024opencap,
  author = {Kakavand, Reza and Ahmadi, Reza and Parsaei, Atousa and Edwards, W. Brent and Komeili, Amin},
  title = {Comparison of Kinematics and Kinetics Between {OpenCap} and a Marker-Based Motion Capture System in Cycling},
  howpublished = {arXiv:2409.03766},
  year = {2024},
  note = {Revised April 2025},
  doi = {10.48550/arXiv.2409.03766}
}

@article{debock2021cyclocross,
  author = {De Bock, Jelle and Verstockt, Steven},
  title = {Video-Based Analysis and Reporting of Riding Behavior in Cyclocross Segments},
  journal = {Sensors},
  volume = {21},
  number = {22},
  pages = {7619},
  year = {2021},
  doi = {10.3390/s21227619}
}

@book{bini2014cycling,
  editor = {Bini, Rodrigo R. and Carpes, Felipe P.},
  title = {Biomechanics of Cycling},
  publisher = {Springer},
  address = {Cham},
  year = {2014},
  doi = {10.1007/978-3-319-05539-8}
}

@article{fonda2012uphill,
  author = {Fonda, Borut and {\v S}arabon, Nejc},
  title = {Biomechanics and Energetics of Uphill Cycling: A Review},
  journal = {Kinesiology},
  volume = {44},
  number = {1},
  pages = {5--17},
  year = {2012},
  url = {https://hrcak.srce.hr/83577}
}

@misc{bikefitia,
  author = {{BikeFit IA}},
  title = {Free Online Bike Fitting: {AI} Cycling Position Analysis},
  howpublished = {Product website. \url{https://www.bikefit-ia.com/en}},
  note = {Accessed 26 September 2026; publication date not specified},
  url = {https://www.bikefit-ia.com/en}
}

@article{engeler2025downhill,
  author = {Engeler, Nadine and Lichtenstein, Eric and Faude, Oliver and Roth, Ralf},
  title = {How Downhill and Uphill Running Interfere Posture and Muscle Activity: A Descriptive Laboratory Study},
  journal = {International Journal of Sports Physical Therapy},
  volume = {20},
  number = {8},
  year = {2025},
  doi = {10.26603/001c.142485}
}
\end{document}